\documentclass[]{bytedance_seed}

\usepackage[toc,page,header]{appendix}

\usepackage{minitoc}
\usepackage[utf8]{inputenc} 
\usepackage[T1]{fontenc}    
\usepackage{hyperref}       
\usepackage{url}            
\usepackage{booktabs}       
\usepackage{amsfonts}       
\usepackage{nicefrac}       
\usepackage{microtype}      
\usepackage{xcolor}         
\usepackage{enumitem}
\usepackage{adjustbox}
\usepackage{framed}
\usepackage{amsmath}
\usepackage{amssymb}
\usepackage{bm}
\usepackage{graphicx}
\usepackage{multirow}
\usepackage[most]{tcolorbox}      
\usepackage{titlesec}             
\usepackage{fontawesome} 
\usepackage{float}
\usepackage{wrapfig}
\usepackage{tikz}

\title{AttentionInfluence: Adopting Attention Head Influence for Weak-to-Strong Pretraining Data Selection}

\author[\dagger]{Kai Hua}
\author[]{Steven Wu}
\author[]{Ge Zhang}
\author[\dagger]{Ke Shen}

\affiliation[]{ByteDance Seed}

\contribution[\dagger]{Corresponding authors}

\abstract{
Recently, there has been growing interest in collecting reasoning-intensive pretraining data to improve LLMs' complex reasoning ability. 
Prior approaches typically rely on supervised classifiers to identify such data, which requires labeling by humans or LLMs, often introducing domain-specific biases.
Due to the attention heads being crucial to in-context reasoning, we propose \textbf{AttentionInfluence}, a simple yet effective, \textbf{training-free} method \textbf{without supervision signal.
}
Our approach enables a \textbf{small pretrained language model} to act as a strong data selector through a simple attention head masking operation. 
Specifically, we identify retrieval heads and compute the loss difference when masking these heads. 
We apply AttentionInfluence to a 1.3B-parameter dense model to conduct data selection on the SmolLM corpus of 241B tokens, and mix the SmolLM corpus with the selected subset comprising 73B tokens to pretrain a 7B-parameter dense model using 1T training tokens and WSD learning rate scheduling. 
Our experimental results demonstrate substantial improvements, ranging from \textbf{1.4pp} to \textbf{3.5pp}, across several knowledge-intensive and reasoning-heavy benchmarks (i.e., MMLU, MMLU-Pro, AGIEval-en, GSM8K, and HumanEval).
This demonstrates an effective \textbf{weak-to-strong} scaling property, with small models improving the final performance of larger models—offering a promising and scalable path for reasoning-centric data selection.
}

\date{\today}
\correspondence{Kai Hua at \email{huakai.dev@bytedance.com}, Ke Shen at \email{shenke@bytedance.com}}

\begin{document}
\maketitle


\section{Introduction}
The identification of high-quality pretraining data has been a key factor enabling Large Language Models' (\textbf{LLMs}) creation.
Commonly recognized high-quality pretraining materials include academic papers (e.g., arXiv), books (e.g., Project Gutenberg), high-quality code (e.g., GitHub), and instruction datasets~\citep{li2024scalingfilter}.
Existing approaches often rely on manually curated high-quality seed data to train classifiers for extracting additional high-quality pretraining data from massive web corpora.
However, as the size and diversity of LLMs' pretraining data requirements continue to grow, these carefully curated classifiers suffer from the high manual effort requirements and relatively low diversity of identified data.
This raises a critical research question: \textit{How can we continue to identify diverse high-quality pretraining data efficiently and scalably?}

\begin{figure*}[!tb]
    \centering
    \includegraphics[width=1\linewidth]{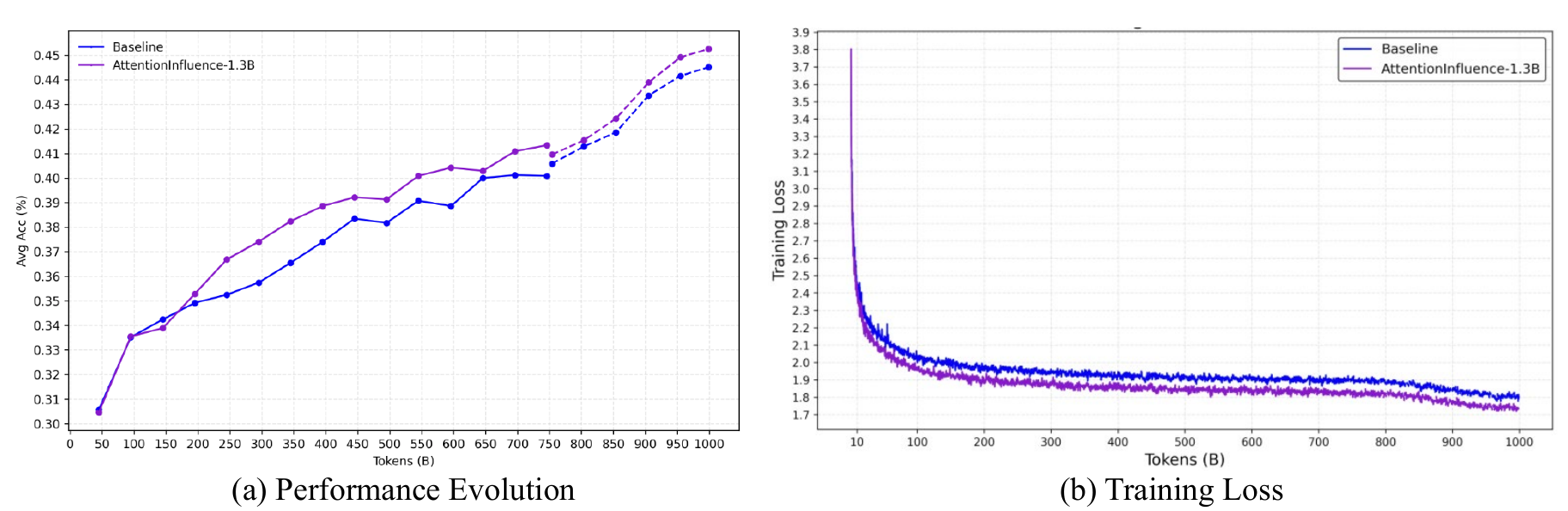}
\caption{
\textbf{(a) Performance evolution on comprehensive benchmark evaluations during pretraining.} The first 750 billion tokens correspond to the pretraining phase, represented by solid lines, while the subsequent 250 billion tokens represent the learning rate annealing phase, represented by dashed lines, using the same dataset.
After around 100 billion tokens, AttentionInfluence-1.3B consistently outperforms the baseline across a wide range of tasks on average, including the annealing phase.
\textbf{(b) Training Loss during pretraining.} AttentionInfluence-1.3B consistently achieves a lower loss than the baseline.
}
    \label{fig:performance_evolution}
\end{figure*}


Current mainstream methods\cite{su2024nemotron} typically use supervised or weakly supervised data to train classifiers to identify high-quality data. 
For instance, LLaMA2\cite{touvron2023llama} uses reference information of Wikipedia documents, which can be seen as weakly supervised data to train a fasttext\cite{joulin2016fasttext} classifier and then recognize Wikipedia-like documents. LLaMA3\cite{grattafiori2024llama} and FineWeb-Edu\cite{penedo2024fineweb} use LLM-generated responses to train a classifier for educational value, which can be regarded as a much sparser form of distillation from a larger LLM(up to 70B dense parameters) than knowledge distillation\cite{hinton2015distilling}. 
While other approaches like DCLM aim to fit user preferences through utilizing signals of user behavior, these methods may introduce potential bias and do harm to diversity\cite{li2024scalingfilter}. 
There are also efforts to train several domain classifiers and combine them for practical use \cite{wettig2025organize}. 
However, we assume that these methods fail to capture the essence of what makes data reasoning-intensive, and as a result, they can be labor-intensive and require significant data engineering efforts. 
Moreover, there exists a risk that the classification results from small models distilled from larger models' responses may not improve the final performance of larger models.


Therefore, we propose \textbf{AttentionInfluence}, which leverages the intrinsic mechanism of existing LLMs' attention heads for pretraining data selection to achieve weak-to-strong generalization.
Existing research suggests that feedforward networks (FFNs) store atomic knowledge\cite{geva2020transformer}, while attention mechanisms execute algorithms and store procedural knowledge\cite{olsson2022context,wu2024retrieval}.
These mechanistic interpretability insights inspire us to hypothesize that the data activating more important attention heads are high-quality and about procedural knowledge.
To be specific, we select the data with a relatively larger loss difference when small pretrained language models process them with and without masking retrieval heads.
Compared with mainstream data selection methods~\citep{li2024scalingfilter,joulin2016fasttext}, our method is training-free and more generalizable.

To validate AttentionInfluence, we adopt a LLaMA2-alike-1.3B pretrained checkpoint for data selection from SmolLM-Corpus. We then pretrain a 7B dense language model---{SmolLM-7B}---as our baseline on the SmolLM-Corpus, a 241B-token curated dataset that already applies strong quality filtering with an education-focused classifier (FineWeb-Edu-Dedup) to retain high-quality data. As shown in (a) of \autoref{fig:performance_evolution}, despite this strong baseline, {AttentionInfluence} still yields consistent improvements, demonstrating its ability to further improve the overall data quality through better data selection beyond existing heuristics or classifiers.
AttentionInfluence shows consistent improvement against SmolLM-7B across a wide range of tasks, demonstrating the effectiveness of the selected data. We further compare AttentionInfluence with a trained classifier (FineWeb-Edu Classifier) and find that it selects data that is more balanced and broadly distributed across content categories, and preferentially favors longer and more comprehensive samples. Despite being entirely supervision-free and training-free, AttentionInfluence also shows strong agreement with classifier-based patterns, validating its reliability and generalizability.

In summary, our key contributions are as follows:
\begin{enumerate}
\item We propose \textbf{{AttentionInfluence}}, a novel framework that leverages intrinsic model behaviors—specifically attention head mechanisms—for \textbf{effective data selection without any supervision signal}.
\item We show that data selected by {AttentionInfluence} is \textbf{high-quality and well-distributed}, yielding consistent improvements in downstream training.
\item We demonstrate that this approach exhibits \textbf{“weak-to-strong” scaling property}, where data selected by smaller models significantly improves the training of larger models, resulting in performance gains without relying on human-labeled data, LLM-generated data, or training any classifiers.
\end{enumerate}

\section{Related Work}
\subsection{Data Selection}
Many works use heuristic filtering rules\cite{rae2021scaling} or perplexity\cite{ankner2024perplexed} of existing LLMs to assess the quality of pretraining data, which makes them training-free. 
Scaling filter\cite{li2024scalingfilter} uses the perplexity difference between two LLMs trained on the same data to evaluate text quality. 
However, when two LLMs trained on the same data are unavailable, training LLMs incurs substantial computational cost.
In contrast, methods that rely on training a model with high-quality labeled data gain more attention owing to their higher accuracy and superior versatility across different data categories. 
For instance, LLaMA2 uses reference information from Wikipedia documents, which can be seen as weak supervision data to train a fasttext\cite{joulin2016fasttext} classifier and then recognize Wikipedia-like documents. LLaMA3\cite{grattafiori2024llama} and FineWeb-Edu\cite{penedo2024fineweb} use LLM-generated responses to train a classifier for educational value, which can be regarded as a much sparser form of distillation from a larger LLM (up to 70B dense parameters) than knowledge distillation\cite{hinton2015distilling}. 
While other approaches like DCLM\cite{li2024datacomp} aim to fit user preferences by utilizing user behavior signals. Some recent and more advanced approaches\cite{wettig2024qurating,zhao2024decoratelm,peng2025dataman} instead train multi-class classifiers to make fine-grained distinctions among various content types and depend on labeled data generated by proprietary commercial large language models such as GPT-3.5-turbo, GPT-4, and GPT-4o. 
There are also efforts to train several domain classifiers and combine them for practical use \cite{wettig2025organize}. 
AttentionInfluence can be seen as a \textbf{training-free} method and is available at minimal cost without any training cost or human-labeled or LLM-labeled data.

\subsection{Data Mixture}
There are also efforts to optimize the data mixture by either relying on human selection or using automatic frameworks.
\citet{ye2024data} propose Data Mixing Laws by introducing a nested prediction framework that combines scaling laws of training steps, model sizes, and data mixtures, enabling efficient optimization of large-scale pretraining data using only small-scale experiments.
\citet{xie2023doremi} propose DoReMi (Domain Reweighting with Minimax Optimization), a method that uses a small proxy model and distributionally robust optimization to automatically learn optimal domain mixture weights for language model pretraining.
\citet{held2025optimizing} conduct data mixing by designing a lightweight method that leverages LLMs to estimate data utility
from small samples, enabling compute-efficient optimization with comparable performance to ablation-based approaches.
\citet{olmo20242} introduce OLMo and uses the existing data mixture designed for a different model family.
REGMIX~\citep{liu2024regmix} is a regression-based framework for optimizing data mixtures in language model pretraining by training small proxy models on diverse mixtures and predicting performance using regression.



\subsection{Mechanistic Interpretability}
Understanding the inner workings of LLMs is crucial for advancing artificial general intelligence safely.
Consequently, studies in mechanistic interpretability \cite{olsson2022context, anthropic_sparse_autoencoder, zheng2024attention, wu2024retrieval, fu2024not, lv2024interpreting} are increasingly being conducted. \citet{olsson2022context} primarily investigates the relationship between certain heads in large language models (LLMs) and their in-context learning capabilities. \citet{anthropic_sparse_autoencoder} extracts a large number of interpretable features from a one-layer transformer with a sparse autoencoder in order to analyze the behavior of the neural network.
\citet{lv2024interpreting} investigates mechanisms for factual recall in Transformer-based LLMs, focusing on attention head extraction, MLP activation, and the collaborative mechanism between attention heads and MLPs. \citet{wu2024retrieval} identifies a class of attention heads, termed retrieval heads, which retrieve relevant information in LLMs. These heads exhibit universal, sparse, and dynamically activated behavior, and they play a crucial role in enabling chain-of-thought reasoning.
\citet{fu2024not} aims to estimate the importance of different attention heads for contextual QA tasks that require both retrieval and reasoning capabilities, enabling efficient head-level KV cache compression for language model inference. 
Inspired by the findings of \citet{wu2024retrieval,qiu2024clongeval}, AttentionInfluence adopts a 
similar and simple proxy task to detect specific important heads, namely the retrieval heads in this paper. 
AttentionInfluence naturally extends the insights from \citet{wu2024retrieval}, broadening their application beyond model analysis and inference acceleration to include effective and efficient data selection.

\subsection{Influence Measure}
\citet{ruis2024procedural} uses influence
functions to recognize pretraining documents important for learning factual knowledge and mathematical reasoning separately. Mirror Influence\cite{ko2024mirrored} realizes an efficient data influence estimation to select high-quality data. MATES\cite{yu2024mates} continuously adapts a data influence model to the evolving data preferences of the pretraining model and then selects the most effective data for the current pretraining progress. Our work is similar to Mirror Influence in that we use data influence estimation to select high-quality data. However, while Mirror Influence requires a high-quality dataset to train a strong reference model and create a model pair with significant differences in capabilities to compute delta loss, our approach uses the attention mechanism to derive a weaker reference model from the base model. This enables us to obtain two models with a significant capability gap and compute delta loss to evaluate data quality.

\section{Preliminary}
\label{sec: Preliminary}
To estimate the impact of each pretraining data sample on LLMs’ intrinsic reasoning and retrieval capabilities, we adapt the retrieval score defined in \citep{wu2024retrieval} and model it as a token-level recall rate based on the attention head behavior.
We denote the current token being generated as $w$ at the decoding step $t$ while decoding the LLM.
We further denote the attention scores of a head as $\boldsymbol{a} \in \mathcal{R}^{|\mathbf{x}|}$ where $x$ represents the vocabulary.
Consequently, $|\mathbf{x}|$ denotes the size of the vocabulary. 
We assume that an attention head $h$ performs a copy-paste operation for corresponding content $\mathbf{k}$, if and only if the following two conditions are met:
\begin{itemize}[left=0pt, align=left]
    \item[] \textbf{Condition 1:} The generated token $w$ appears in the corresponding content $k$:
    \begin{align}
        w \in \mathbf{k}
    \end{align}

    \item[] \textbf{Condition 2:} The token $w$ receives the highest attention score among all positions visible to the current query token in this head:
    \begin{equation}
        j \in \mathbf{i}_q,\;\text{where $\mathbf{i}_q = \{ j \mid j < t \}$ is the set of positions visible at decoding step $t$}
    \end{equation}
    \begin{equation}
        j = \arg\max(\boldsymbol{a}),\;\mathbf{x}_j = w
    \end{equation}
\end{itemize}

Let $\mathbf{g_h}$ denote the set containing all tokens copied and pasted by a given head $h$, we define:
\begin{equation}
    \text{Retrieval score for head}\;\; h = \frac{|\mathbf{g_h} \cap \mathbf{k} |}{|\mathbf{k}|}
\end{equation}

\section{Method}
\citet{lin2024rho} demonstrate that a well-trained reference model can serve as a proxy to fit the desired data distribution of the LLM pretraining by comparing the data loss gap between the base model and the reference model. By comparing the token-level data loss gap between the base model and the reference model, they can identify important tokens that align better with the target distribution. Inspired by recent work~\citet{lin2024rho, ko2024mirrored}, we propose \textbf{AttentionInfluence} to select high-quality pretraining data based on the data loss gap from a <weak model, strong model> pair. However, while existing approaches~\citep{lin2024rho,ko2024mirrored} focus on building a stronger reference model as the \textit{strong model}, AttentionInfluence points out that it is cheaper and more controllable to degrade the base model to a weaker version, thus constructing a <weak model, strong model> pair.

\begin{figure}[H]
    \centering
    \includegraphics[width=0.99\linewidth]{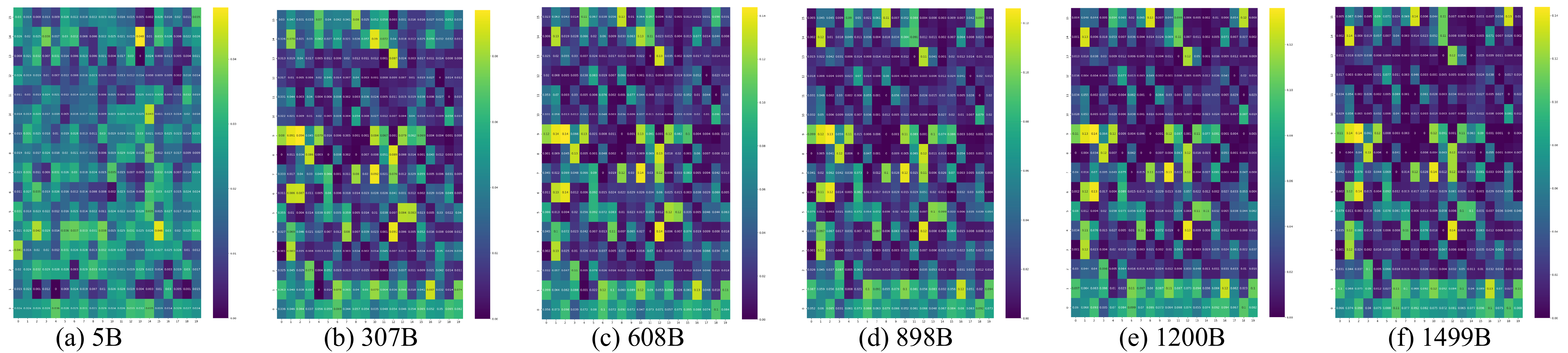}
    \caption{The evolution of retrieval heads in a 1.3B dense model.}
    \label{fig:evolution_of_1.3B}
\end{figure}

Existing studies~\citep{olsson2022context, wu2024retrieval} point out that specific attention heads (i.e., \textbf{retrieval heads}) plays a critical role in LLMs' in-context learning, retrieval, and reasoning capabilities.
We find that the language model's retrieval heads emerge early in training, gradually strengthen, and eventually become entrenched in the middle to late stages, playing a crucial role in the model's performance, as shown in \autoref{fig:evolution_of_1.3B}.
Inspired by this insight, AttentionInfluence identifies the specific attention heads that are important for targeted LLM capabilities and obtains a degraded reference model by disabling them.
Then, AttentionInfluence selects high-quality pretraining data based on the sample-level data loss gap from the constructed <weak model, strong model> pair.

We detail the AttentionInfluence method in the following section.

\subsection{Detecting Specific Important Heads}
\label{sec:detect}
In this work, we detect the retrieval heads as specifically important heads for reasoning, because \citet{wu2024retrieval} reveals that retrieval heads are extremely relevant to LLMs' retrieval and reasoning ability.

Inspired by the Key-Passage Retrieval evaluation task proposed in CLongEval\cite{qiu2024clongeval}, we adopt a similar and simple proxy task to evaluate the retrieval ability of LLMs in a controlled setting, and identify attention heads that are strongly associated with retrieval and reasoning. To this end, we construct a synthetic 
test dataset consisting of 800 samples. Each sample is formatted as a 3-shot retrieval task in natural language, consisting of a \texttt{context}, three in-context demonstrations, and a query \texttt{hash\_key}. The sample template is detailed in \autoref{sec:synthetic_test_sample}. Each \texttt{context} is a JSON object with $k$ key-value pairs, where each key is a randomly generated 32-character alphanumeric string (\texttt{hash\_key}), and each value (\texttt{text\_val})\footnote{Each \texttt{text\_val} is capped at a maximum of 30 tokens.} is a natural language sentence sampled from a corpus of web documents. The task requires the model to retrieve the \texttt{text\_val} from the \texttt{context} and output the \texttt{text\_val} corresponding to the given query \texttt{hash\_key}. The inclusion of three in-context demonstrations (i.e., 3-shot) is designed to simulate a few-shot learning scenario and help the model understand the task. Considering the context length limitation of existing pretrained models, we constrain the total length of each test sample---including both the input prompt and the answer—to be close to but not exceeding 4,096 tokens.






\begin{figure*}[!tb]
    \centering
    \includegraphics[width=0.7\linewidth]{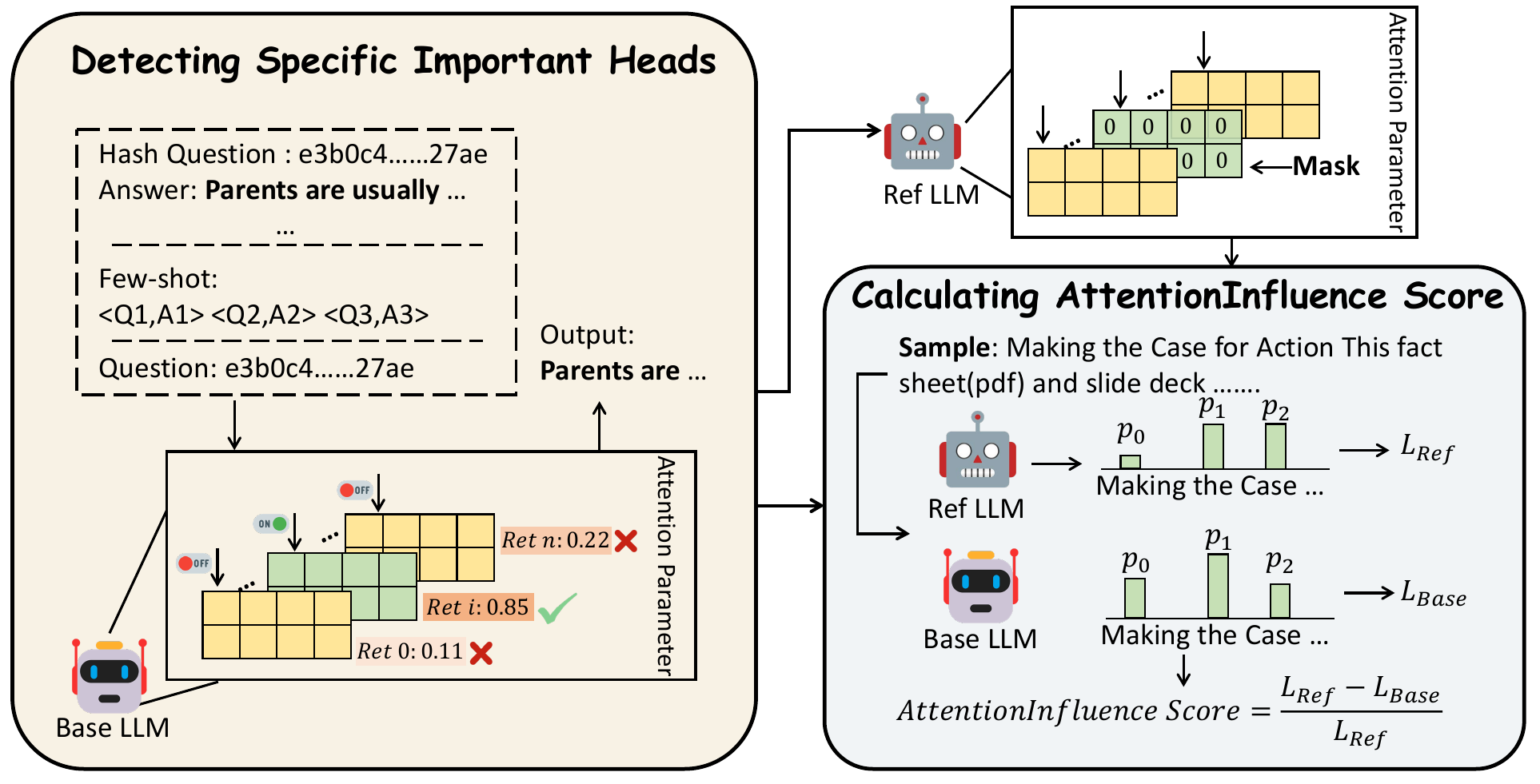}
    \caption{The illustration of AttentionInfluence.}
    \label{fig:method}
\end{figure*}


Next, we compute retrieval scores for each attention head across test samples, as described in Section~\ref{sec: Preliminary}. In this work, we use a 1.3B-parameter model based on the LLaMA2-alike architecture as the small pretrained language model. We use the average score as the head’s final retrieval score and sort them by it. Referring to \citet{wu2024retrieval}, 
we select the heads ranked in the top 5\% as specifically important heads.



\subsection{Calculating AttentionInfluence Score}
We obtain a reference model by masking the important heads of the base model detected in the first phase, and compute the AttentionInfluence score based on the base model and reference model.
For details on the masking operation, refer to \autoref{sec:Masking_Operation}.
First, we use the base model to compute the mean token-level cross-entropy loss ($\mathcal{L}_{\mathrm{base}}$)
 of each sample in the corpus.
Subsequently, we compute the corresponding loss ($\mathcal{L}_{\mathrm{ref}}$) using the reference model.
Finally, we use the relative delta between $\mathcal{L}_{\mathrm{base}}$ and  $\mathcal{L}_{\mathrm{ref}}$ as an AttentionInfluence Score to quantify the reasoning intensity of each sample, which can be denoted as:
\begin{equation}
\text{AttentionInfluence Score} = \frac{\mathcal{L}_{\mathrm{ref}} - \mathcal{L}_{\mathrm{base}}}{\mathcal{L}_{\mathrm{base}}}
\end{equation}

Since the loss of a language model for data from different domains (e.g., general/math/code) cannot be directly compared due to significant distribution differences, we restrict the AttentionInfluence Score to be compared only within the same domain (e.g., general/math/code). 
We consider that a higher AttentionInfluence Score indicates a higher reasoning intensity of the sample.

\begin{table}[htbp]
\centering
\resizebox{0.85\textwidth}{!}{%
\begin{tabular}{llccccccccc}
\toprule
Model  &	\#Tokens & Avg. & \multicolumn{6}{c}{Metrics}
\\
\bottomrule \bottomrule
\multirow{6}{*}{SmolLM-1.7B}  	& \multirow{6}{*}{1T} & \multirow{6}{*}{-}  &   

ARC-C &	 ARC-E &	 ARC(C+E) &	 Wino. 	& Hella. &	 CSQA & OpenBookQA  \\
 & & &  -&	 - & 59.95 &	54.70&	62.83&	38.00 & 42.60 \\
 & & & PIQA& 	 TriviaQA&	 MMLU   &  MMLU-Pro &	AGIEval-en & RACE 	 &DROP 	   \\
&  &   & 75.90  & 13.14 & 39.35	&10.92	&-  & - &	 -   \\
&  &  & BBH &	 GSM8K& 	 MATH &	 HumanEval & C-Eval & GPQA  \\
& &  & - 	 & \ 4.62 	& -	&-  & -	&- \\

\midrule\midrule

\multirow{6}{*}{Baseline w/o LRD}   	& \multirow{6}{*}{746B}  &   \multirow{6}{*}{40.09}  &   

ARC-C &	 ARC-E &	 ARC(C+E) &	 Wino. 	& Hella. &	 CSQA & OpenBookQA  \\
 & & & 55.89&	81.69	&68.79	&66.22&	71.79	& {49.14} & {45.40}	 \\
 & & & PIQA& 	 TriviaQA&	 MMLU   &  MMLU-Pro &	AGIEval-en & RACE 	 &DROP 	   \\
 & &   & 79.27	&45.57	&41.76 & 13.80&	21.10 & 40.67	&31.71	  \\
&  &  & BBH &	 GSM8K& 	 MATH &	 HumanEval & C-Eval & GPQA  \\
& &  & 30.35	&12.89	& \ 6.15	&20.70 & 26.08	& 21.93

 \\
 \midrule\midrule

\multirow{6}{*}{Ours w/o LRD} & \multirow{6}{*}{746B}  & \multirow{6}{*}{\colorbox{green!15}{41.34}}  &    

ARC-C &	 ARC-E &	 ARC(C+E) &	 Wino. 	& Hella. &	 CSQA & OpenBookQA  \\
 & & & \colorbox{green!15}{56.66} &	 \colorbox{green!15}{82.03} 	& \colorbox{green!15}{{69.35}} &	 {65.43}&	 \colorbox{green!15}{71.90} &	 \colorbox{green!15}{53.48} & {43.60} \\
 & & & PIQA& 	 TriviaQA&	 MMLU   &  MMLU-Pro &	AGIEval-en & RACE 	 &DROP 	   \\
 & &  & {77.58} &	 \colorbox{green!15}{45.68}	 &\colorbox{green!15}{{45.10}}  & \colorbox{green!15}{{17.19}} &	22.50	& \colorbox{green!15}{41.72} 	& \colorbox{green!15}{{32.03}} \\
&  &  & BBH &	 GSM8K& 	 MATH &	 HumanEval & C-Eval & GPQA  \\
 & &   & \colorbox{green!15}{{33.99}} 	& \colorbox{green!15}{15.77}&	 \ \colorbox{green!15}{7.25}  	 &\colorbox{green!15}{19.85} & 28.45& 25.18

 \\
 \midrule\midrule

\multirow{6}{*}{Baseline w/ LRD}  	& \multirow{6}{*}{1T}  & \multirow{6}{*}{44.51}  &   

ARC-C &	 ARC-E &	 ARC(C+E) &	 Wino. 	& Hella. &	 CSQA & OpenBookQA  \\
 & & & 58.79& 83.92	&71.36	&70.24	&75.63	& {59.62} & {48.00} \\
 & & & PIQA& 	 TriviaQA&	 MMLU   &  MMLU-Pro &	AGIEval-en & RACE 	 &DROP 	   \\
 & &  &  80.63	&51.07	&50.05 &19.36&	24.50& 41.15	&36.09	  \\
&  &  & BBH &	 GSM8K& 	 MATH &	 HumanEval & C-Eval & GPQA  \\
& &  & 35.36&	21.00	& \ 9.00	&23.02 	& 33.80 & 24.77

 \\
 \midrule\midrule

\multirow{6}{*}{Ours w/ LRD }  	& \multirow{6}{*}{1T}  & \multirow{6}{*}{\colorbox{green!15}{45.26}}  &

ARC-C &	 ARC-E &	 ARC(C+E) &	 Wino. 	& Hella. &	 CSQA & OpenBookQA  \\
 & & & \colorbox{green!15}{59.98} 	& \colorbox{green!15}{84.26} 	& \colorbox{green!15}{{72.12}} 	& {68.03} 	 &{75.49} &	 \colorbox{green!15}{61.59} & {46.60}  \\
 & & & PIQA& 	 TriviaQA&	 MMLU   &  MMLU-Pro &	AGIEval-en & RACE 	 &DROP 	   \\
&  &   & {79.54} &	 \colorbox{green!15}{51.20}	 &\colorbox{green!15}{{51.48}} & \colorbox{green!15}{{22.03}} &	\colorbox{green!15}{{26.30}}  & \colorbox{green!15}{42.30} 	& \colorbox{green!15}{36.52}   \\
&  &  & BBH &	 GSM8K& 	 MATH &	 HumanEval & C-Eval & GPQA  \\
& & & \colorbox{green!15}{36.22} &	 \colorbox{green!15}{23.73} &	\ \colorbox{green!15}{10.80}& 	 \colorbox{green!15}{26.55} & 33.06& 24.26	 
 \\

\bottomrule
\end{tabular}%
}
\caption{The main results on various benchmarks. The LRD denotes learning rate decay. 
}
\label{tab:main_result}
\end{table}

\section{Experiments and Results}
In this section, we present experimental analyses to validate the effectiveness of reasoning-intensive data selected by AttentionInfluence.

\label{evaluation}
\subsection{Experimental Details}

We apply AttentionInfluence to a \textbf{LLaMA2-alike-1.3B} pretrained model to rank the SmolLM-Corpus dataset\footnote{\url{https://github.com/huggingface/smollm/tree/main/text/pretraining}}~\citep{benallal2024smollmcorpus}. The specifications of the model are described in \autoref{sec:models_used_by_attention_influence}.
Specifically, we select the top 20\% of samples based on the AttentionInfluence score, yielding approximately \textbf{73.1B} reasoning-intensive tokens for pretraining.

To evaluate the effectiveness of AttentionInfluence, we pretrain a \textbf{7B} dense model using a combination of the full SmolLM-Corpus and the selected 73.1B tokens.
For comparison, we pretrain another model of identical architecture and size using only the SmolLM-Corpus, serving as the baseline.
The model architecture follows that of LLaMA2, and detailed hyperparameters are listed in \autoref{tab:model_hyperparams}.
Further information on the SmolLM-Corpus dataset and pretraining configurations is provided in \autoref{Sec: Experiment Setting}.

Following \citet{grattafiori2024llama}, we adopt a comprehensive set of benchmark evaluations across \textbf{four} major categories under the few-shot setting to holistically compare our model with the baseline: \textbf{1) Aggregate Benchmarks}, including AGIEval-en~\cite{zhong2023agieval}, MMLU~\cite{hendrycks2020measuring}, MMLU-Pro~\cite{wang2024mmlu}, GPQA~\cite{rein2023gpqagraduatelevelgoogleproofqa}, and C-Eval~\cite{huang2023c}; \textbf{2) Mathematics, Code, and Reasoning}, comprising GSM8K~\cite{cobbe2021training}, MATH~\cite{hendrycks2021measuring}, HumanEval~\cite{chen2021evaluating}, ARC Challenge~\cite{clark2018think}, DROP~\cite{dua2019drop}, and BBH~\cite{suzgun2022challenging}; \textbf{3) Commonsense Reasoning and Understanding}, including HellaSwag~\cite{zellers2019hellaswag}, ARC-Easy~\cite{clark2018think}, WinoGrande~\cite{sakaguchi2021winogrande}, CommonSenseQA~\cite{talmor2018commonsenseqa}, PiQA~\cite{bisk2020piqa}, OpenBookQA~\cite{mihaylov2018can}, and TriviaQA~\cite{joshi2017triviaqa}; and \textbf{4) Reading Comprehension}, represented by RACE~\cite{lai2017racelargescalereadingcomprehension}. 
Details of the evaluation setup are provided in \autoref{Sec: Experiment Setting}.


\subsection{Results}
\textbf{1) Aggregate Benchmarks:} On challenging aggregate benchmarks such as MMLU, MMLU-Pro, and AGIEval-en, AttentionInfluence consistently outperforms the baseline, indicating stronger general knowledge and reasoning capabilities. These improvements—\textbf{+1.4pp} on MMLU, \textbf{+2.7pp} on MMLU-Pro, and \textbf{+1.8pp} on AGIEval-en—underscore the effectiveness of AttentionInfluence in identifying a diverse distribution of pretraining data that supports both \textbf{comprehensive knowledge acquisition} and \textbf{reasoning-intensive learning}.

\textbf{2) Math, Code, and Reasoning:} AttentionInfluence yields substantial improvements on complex multi-step reasoning tasks such as GSM8K (\textbf{+2.7pp}), HumanEval (\textbf{+3.5pp}), and BBH (\textbf{+0.9pp}), suggesting that the selected data distribution better facilitates problem-solving and advanced reasoning. Additional gains on ARC-Challenge, DROP, and MATH further demonstrate that \textbf{AttentionInfluence enhances reasoning generalization across a wide range of tasks}.

\textbf{3) Commonsense Reasoning and Understanding:} On benchmarks including CSQA, PiQA, and OpenBookQA, AttentionInfluence achieves competitive or superior results. Notably, its performance on ARC-Easy and TriviaQA indicates that AttentionInfluence maintains strong results on factual and commonsense tasks, despite being primarily designed to select reasoning-intensive data.

\textbf{4) Reading Comprehension:} On RACE, AttentionInfluence surpasses the baseline by \textbf{+1.2pp}, reflecting enhanced discourse-level understanding and reasoning capabilities.


\textbf{Performance Evolution During Pretraining:}
We evaluate the training dynamics of AttentionInfluence-1.3B on the specified tasks. As shown in \autoref{fig:performance_evolution}, our method consistently outperforms the baseline throughout the pretraining process. 
Full results for all evaluation tasks are provided in \autoref{fig:performance_evolution_full_1} and \autoref{fig:performance_evolution_full_2}.
The performance gap emerges early—well before reaching 100B tokens—and remains stable over time. After approximately 100 billion tokens, AttentionInfluence-1.3B demonstrates a clear and consistent advantage over the baseline, on average, across multiple tasks. These improvements persist throughout all training phases, including both before and after the learning rate decay (LRD). Although the performance margin slightly narrows following the LRD, this effect primarily results from saturation as training approaches 1T tokens on the 7B model, which is trained on the SmolLM-Corpus containing only 241B tokens. Nonetheless, AttentionInfluence maintains a stable advantage without requiring any additional supervision signals.
Moreover, on benchmarks that primarily require simple factual knowledge, the performance of LLMs trained with AttentionInfluence is comparable to that of the baseline (see \autoref{fig:performance_evolution_full_1}).
In contrast, on reasoning-intensive benchmarks, our models achieve significantly better results than the baseline (see \autoref{fig:performance_evolution_full_2}).
These findings demonstrate that AttentionInfluence is effective at selecting data with higher reasoning intensity.

\textbf{Mirror Effects in AttentionInfluence:}
For tasks with performance gains---such as MMLU, MMLU-Pro, AGIEval-en, DROP, BBH and GSM8K---we observe that masking retrieval heads in the pretrained 1.3B model leads to a significant performance drop (see \autoref{appd:different_heads_influence} for details). This suggests a mirror effect: when the performance of the 1.3B model significantly degrades on certain tasks due to masking the specific important heads, the data selected by AttentionInfluence-1.3B tends to improve performance on these tasks when used to train a 7B model. This observation supports the insight discussed in \autoref{sec:detect}, demonstrating the interpretability of AttentionInfluence and its predictive power in identifying evaluation metrics likely to show improvement before any training.

\textbf{Increasing Parameter Size of AttentionInfluce:}
Furthermore, as shown in \autoref{tab:1.3B_7B_difficult_benchs}, LLMs pretrained on data selected by AttentionInfluence-7B exhibit superior performance on these challenging knowledge-intensive and reasoning-intensive benchmarks. This indicates that increasing the model size in AttentionInfluence enables the selection of samples with higher reasoning intensity. The comparison details between the 1.3B and 7B methods are provided in \autoref{tab:abalation_result} of \autoref{sec:extra_exp}.

\textbf{In conclusion:} These results validate that \textbf{AttentionInfluence effectively identifies high-quality pretraining data that enhances the knowledge and reasoning capabilities of LLMs}, yielding particularly notable gains on benchmarks requiring comprehensive knowledge and complex reasoning, as shown in Table~\ref{tab:main_result}.
Furthermore, AttentionInfluence can be combined with the FineWeb-Edu Classifier to achieve comprehensive improvements in LLM performance on tasks that require either simple factual knowledge, advanced reasoning, or both.

\begin{table}[htbp]
\centering
\resizebox{0.5\textwidth}{!}{%
\begin{tabular}{lccccccc}
\toprule
Model & MMLU& GPQA& MATH&  C-Eval & AGIEval-en & BBH \\
\midrule
1.3B & 51.48 &  24.26 & 10.80 &   33.06 & 26.30 & 36.22  \\
7B	& 53.18  &  24.87 & 11.75  &   36.85 & 26.85 & 36.80\\

\bottomrule
\end{tabular}%
}
\caption{Comparison of results trained with AttentionInfluence-1.3B/7B on relatively difficult benchmarks.}
\label{tab:1.3B_7B_difficult_benchs}
\end{table}

\section{Discussion}
\subsection{Reliability of AttentionInfluence}
\label{sec: LLMAsJudge}

\begin{table}[htbp]
\centering
\resizebox{0.85\textwidth}{!}{%
\begin{tabular}{lcccccc}
\toprule
\multirow{2}{*}{Domain} & \multicolumn{3}{c}{FineWeb-Edu Classifier} & \multicolumn{3}{c}{AttentionInfluence}  \\
 \cmidrule(lr){2-4} \cmidrule(lr){5-7}
& Edu Score & Reasoning Score	& Token Len & Edu Score & Reasoning Score	& Token Len
\\
\midrule
FineWeb-Edu-Dedup &	0.99 &	0.52 	&1610.12 	&0.99& 	0.49 &	1629.73 \\
Cosmopedia-V2	&1.0 	&0.87 &	825.46 	&1.0 &	0.80 &	893.805 \\
Python-Edu	& 0.98 &	0.76 &	414.15 &	0.98 &	0.87 &	820.71 \\
OpenWebMath&	0.99 &	0.52 &	1022.855 & 	0.96 &	0.88 &	2255.575 \\

\bottomrule
\end{tabular}%
}
\caption{The quality score of the data selected by AttentionInfluence and FineWeb-Edu Classifier.}
\label{tab:quality_score}
\end{table}

To validate the effectiveness of AttentionInfluence, we design two metrics—\textbf{Education Score} and \textbf{Reasoning Score}—to quantify the quality of the selected data. Specifically, we randomly sample 200 examples from the top 20\% ranked by AttentionInfluence and the FineWeb-Edu classifier, respectively, and employ GPT-4o as the evaluator. The detailed scoring criteria and prompt design for both metrics are provided in \autoref{appd:LLM-As-A-Judge}.

As shown in \autoref{tab:quality_score}, both AttentionInfluence and FineWeb-Edu classifier yield comparable scores on education-related content. However, AttentionInfluence achieves substantially higher scores in reasoning, indicating that \textbf{the samples selected by AttentionInfluence exhibit greater reasoning intensity}.

Additionally, we analyze the length of the selected samples. In the Python-Edu and OpenWebMath domains, AttentionInfluence selects samples with an average length nearly twice that of those selected by the FineWeb-Edu classifier. A qualitative inspection of these samples (see \autoref{sec:case}) reveals that, in the Python-Edu domain, AttentionInfluence prefers documents containing not only more complex code but also richer textual context, such as detailed problem statements. In the OpenWebMath domain, the selected samples demonstrate more elaborate formula-based reasoning. \textbf{These findings suggest that AttentionInfluence effectively identifies data with more comprehensive and complex reasoning structures.}

\subsection{Diversity of Selected Data by AttentionInfluence}
\subsubsection{Word Frequency Analysis} 

\begin{table}[htbp]
\centering
\resizebox{0.79\textwidth}{!}{%
\begin{tabular}{llcccc}
\toprule
\multirow{2}{*}{Ranking (\%)} &  \multirow{2}{*}{Static Method} & \multicolumn{4}{c}{Data Source} \\
 \cmidrule(lr){3-6}
& & FineWeb-Edu-Dedup	& Cosmopedia-v2 & Python-Edu	& OpenWebMath
\\
\midrule
\multirow{2}{*}{10} & TF&	0.84 &0.73 &	0.29 	&0.57 	 \\
&TF-IDF&	0.82 &	0.72 &	0.38 	&0.52  
  \\ 
\hline
\multirow{2}{*}{20} & TF&	0.88 & 	0.81	&0.41 &	0.67 
   \\
&TF-IDF&	0.87 &0.80 & 	0.43 	&0.63 	 
\\
\hline
\multirow{2}{*}{50} & TF	&0.95 &	0.91	&0.67 &	0.79  
 \\
 &TF-IDF	&0.92 &	0.90 &0.66 &	0.78  

\\
\bottomrule
\end{tabular}%
}
\caption{Word overlap by ranking threshold and frequency-based statistical method}
\label{tab:wordoevrlap}
\end{table}

We separately select the top 10\%, 20\%, and 50\% of samples ranked by AttentionInfluence and the FineWeb-Edu classifier, and compute the overlap of high-frequency words using multiple statistical approaches.

As shown in \autoref{tab:wordoevrlap}, we derive several key insights:
1) AttentionInfluence exhibits a high degree of overlap with the FineWeb-Edu Classifier, highlighting the \textbf{reliability of the samples selected by AttentionInfluence}.
2) \textbf{AttentionInfluence and the FineWeb-Edu Classifier demonstrate a degree of complementarity}. 
We observe notable domain-specific variations. Specifically, in the FineWeb-Edu-Dedup and Cosmopedia-v2 domains, the overlap exceeds 70\%, whereas in the Python-Edu and OpenWebMath domains, it falls below 60\%.
To further examine the differences between AttentionInfluence and FineWeb-Edu Classifier in specific domains, we sample representative examples from the Python-Edu and OpenWebMath domains, as shown in \autoref{sec:case}. 
These cases reveal that although the two methods display different preferences across domains, both yield reasonable selections."


As shown in \autoref{tab:wordo_un_evrlap} of \autoref{appendix: High Frequency Words}, \textbf{AttentionInfluence places greater emphasis on method-related terminology, while FineWeb-Edu Classifier is more sensitive to numerical expressions}. We identify two distinctive high-frequency terms: \textit{“19th”} from subset selected by FineWeb-Edu Classifier and \textit{“sklearn”} from AttentionInfluence's subset. We then retrieve representative documents from the original corpus containing these terms. The sample containing \textit{“19th”} is related to historical topics, whereas the one with \textit{“sklearn”} discusses K-Nearest Neighbors Classifier and Hyperparameter Tuning.  
\textbf{This suggests that AttentionInfluence prefers samples containing hands-on coding or procedural mathematical reasoning.}

\subsubsection{Clustering-Based Distribution Analysis}

\begin{figure*}[!tb]
    \centering
    \includegraphics[width=0.8\linewidth]{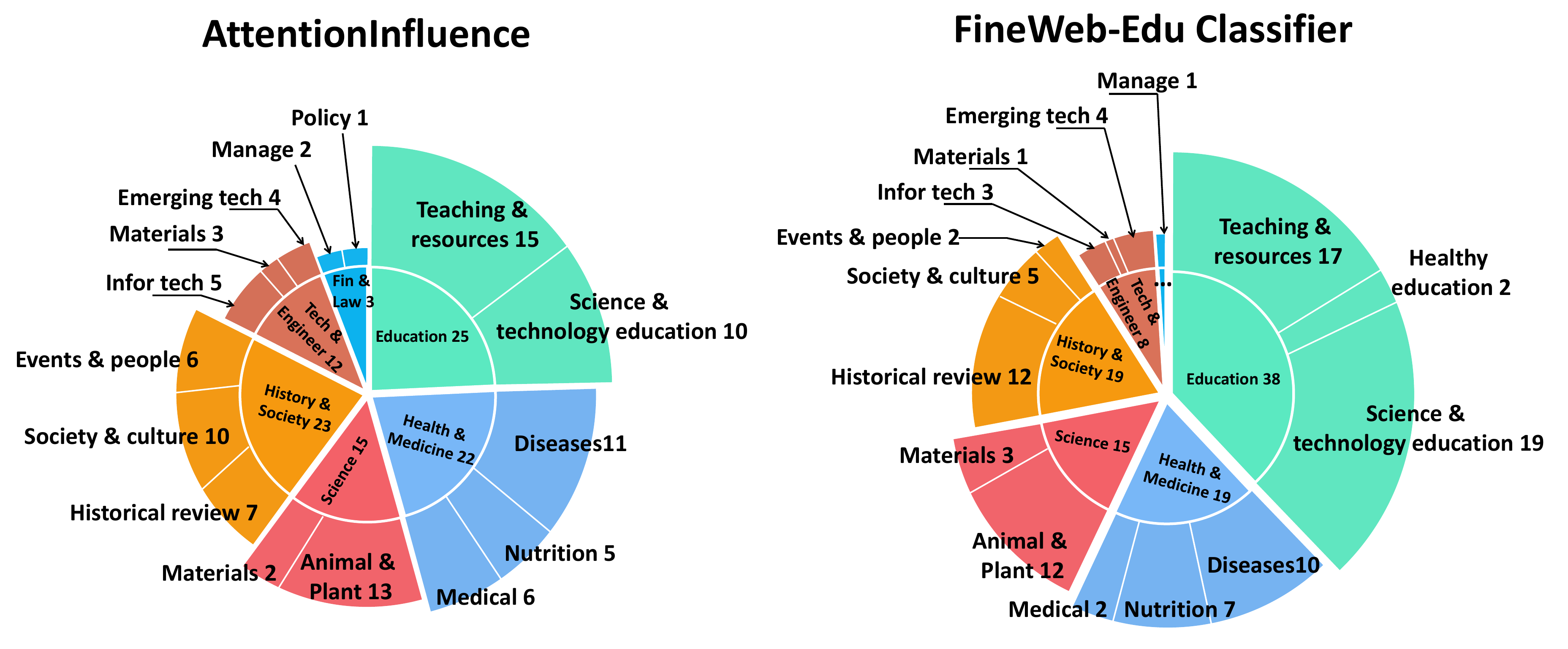}
    \caption{The statistics of clustering. The left is the clustering result of AttentionInfluence, the right part is that of FineWeb-Edu Classifier.}
    \label{fig:ClusteringStatics}
\end{figure*}

To better understand the distribution of samples selected by different methods (i.e., AttentionInfluence and the FineWeb-Edu Classifier), we cluster the selected subsets and employ GPT-4o to annotate the resulting clusters.
The clustering procedure is detailed in \autoref{sec: Details of Clustering}.

We derive the following insights:

\textbf{1) AttentionInfluence produces a more balanced distribution across data categories.}
As illustrated in \autoref{fig:ClusteringStatics}, both methods achieve broad coverage of the top-level categories. However, the distribution resulting from AttentionInfluence is noticeably more balanced.

\textbf{2) AttentionInfluence selects a highly diverse set of samples.}
We examine two clusters from the AttentionInfluence subset that exhibit large embedding distances. As demonstrated by the examples from the \textit{Health Guidelines \& Nutrition} and \textit{Information Technology} clusters in \autoref{Sec:ClusteringCase}, the selected samples differ substantially in both content and style. This lack of semantic overlap underscores the effectiveness of the clustering and enhances the interpretability of the annotated categories.

\subsubsection{The Visualization of Data Distribution}
To provide an intuitive illustration of the relationship between the two selection methods, we apply Principal Component Analysis (PCA) to reduce the dimensionality of the document embeddings and visualize their distributions in two-dimensional space.

\begin{wrapfigure}{r}{0.43\textwidth}  
  \vspace {-0.1cm}
  \centering
  \includegraphics[width=0.4\textwidth]{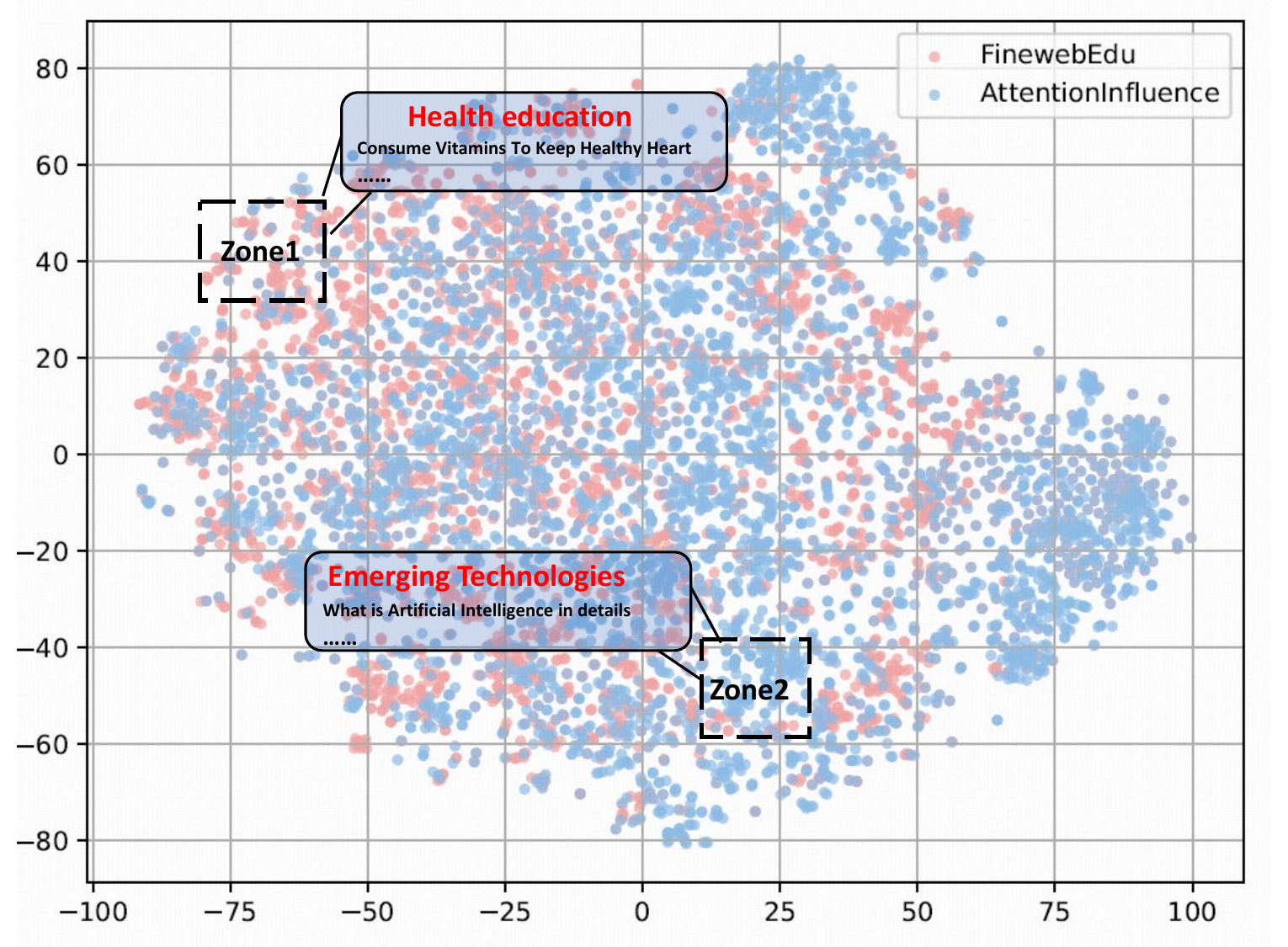}
  \caption{Visualization of data selected by AttentionInfluence and FineWeb-Edu Classifier.}
  \label{fig:embedding_analysis}
\end{wrapfigure}


As shown in \autoref{fig:embedding_analysis}, AttentionInfluence selects samples with broader and more balanced coverage. \textbf{By directly leveraging the attention mechanisms of pretrained language models, it facilitates more effective selection of general and diverse training data than the FineWeb-Edu classifier.}

\textbf{In addition, the selected samples from the two methods exhibit complementary coverage}. 
We further examine the distinctive regions identified by AttentionInfluence and FineWeb-Edu Classifier. For example, the samples in Zone1 are related to Health Education, while most samples in Zone2 fall under the theme of Emerging Technologies. This suggests that the samples selected by the two methods can be complementary. How to effectively integrate the strengths of both selection strategies could be a promising direction for future exploration.


\subsection{Scalability of AttentionInfluence}
\begin{table}[htbp]
\centering
\resizebox{0.85\textwidth}{!}{%
\begin{tabular}{lcccccc}
\toprule
\multirow{2}{*}{Domain} & \multicolumn{3}{c}{1.3B} & \multicolumn{3}{c}{7B}  \\
 \cmidrule(lr){2-4} \cmidrule(lr){5-7}
& Edu Score & Reasoning Score	& Token Len & Edu Score & Reasoning Score	& Token Len
\\
\midrule
FineWeb-Edu-Dedup 
 	&0.99 & 	0.49 &	1895.7 &	0.97  &	0.58 	& 3488.8 \\
Cosmopedia-V2 &1.0 &	0.80 &	2774.6 &1.0 	&0.82 & 2984.1 \\
Python-Edu	
 &	0.97 &	0.87 &	909.3 &0.98 &	0.91 &	1657.2 \\
OpenWebMath
& 	0.96 &	0.88 &	2138.6 &	0.96 &	0.93 &	5550.4\\

\bottomrule
\end{tabular}%
}
\caption{The quality score of the data selected by AttentionInfluence using 1.3B and 7B models, respectively.}
\label{tab:quality_score_1.3_7B}
\end{table}


We compare the samples selected by the AttentionInfluence method using 1.3B and 7B pretrained language models.
We obtain the following insights:


\textbf{\textbf{AttentionInfluence based on a larger LLM selects higher quality data.}}
Similar to the setting in the section \ref{sec: LLMAsJudge}, we use GPT4o to evaluate selected samples.
As shown in \autoref{tab:quality_score_1.3_7B}, across all domains, the samples selected by the 7B model exhibit high edu scores that are comparable to those selected by the 1.3B model, with a slight overall advantage. Regarding reasoning scores, the 7B model consistently outperforms the 1.3B model across all four domains, achieving a particularly notable improvement of 9\% in the FineWeb-Edu-Dedup domain. These results suggest that larger models are more effective at identifying reasoning-intensive samples.

\begin{wrapfigure}{r}{0\textwidth}  
  \vspace {-1cm}
  \centering
  \includegraphics[width=0.4\textwidth]{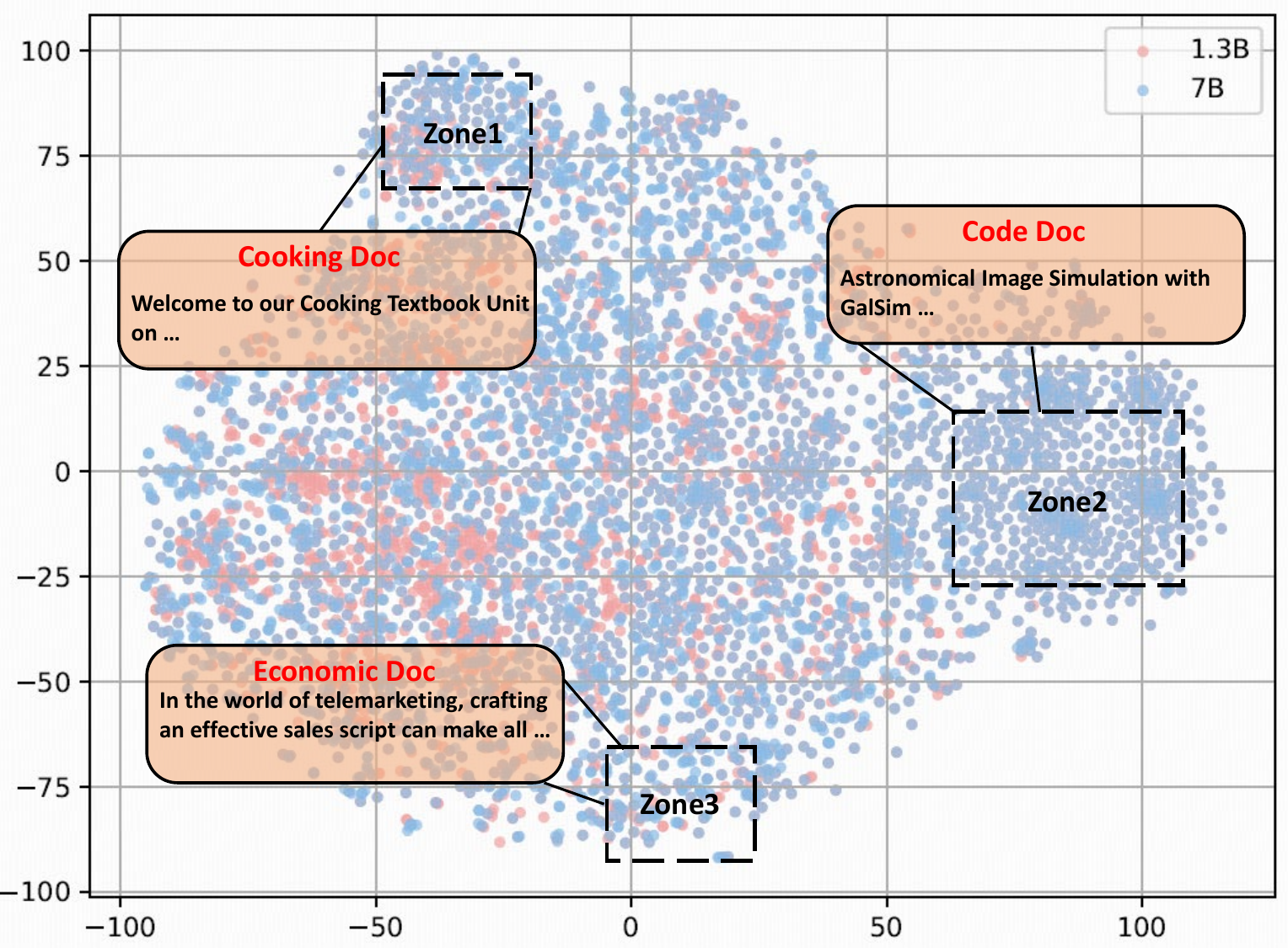}
  \caption{The visualization of the samples selected by AttentionInfluence using 1.3B and 7B models.}
  \label{fig:visualization_1.5_7B}
\end{wrapfigure}
\textbf{\textbf{AttentionInfluence based on a larger LLM is more generalizable.}} As shown in the \autoref{fig:visualization_1.5_7B}, we compare the sample distributions selected by the 1.3B and 7B models. 
We observe that the samples selected by the 7B model are more widely distributed in the space, covering many areas that the 1.3B model fails to reach. 
Notably, regions underrepresented by the 1.3B model are densely populated with specific categories of samples, which are predominantly captured by the 7B model.
For instance, Zone1 corresponds to cooking, Zone2 is related to code, and Zone3 mainly focuses on the economy. This suggests that even without additional training, the samples selected by larger models are more balanced and diverse, capturing a broader range of information. Moreover, we also trained a 7B model on the data selected by AttentionInfluence-7B. As shown in the appendix (see \autoref{tab:abalation_result} and \autoref{fig:performance_evolution_scaling}), this model achieves better performance than AttentionInfluence-1.3B in the middle and later stages of training, where the average accuracy excludes the result of the C-Eval\cite{huang2023c} evaluation task. However, the gap narrows during the final learning rate annealing phase, which is likely due to saturation in the SmolLM corpus and training setup referring to the comparisons with SmolLM\cite{benallal2024smollmcorpus} and SmolLM2\cite{allal2025smollm2}. Importantly, the selected evaluation benchmarks may not fully capture the generalization benefits of AttentionInfluence-7B. For example, while the SmolLM corpus is predominantly English with minimal Chinese contents, we observe that AttentionInfluence-7B significantly outperforms AttentionInfluence-1.3B on the Chinese C-Eval benchmark which is shown in \autoref{fig:performance_evolution_full_2}, reflecting a broader and more robust generalization ability that remains underexplored under the current evaluation settings.

\section{Conclusion}
In this paper, we propose AttentionInfluence, a training-free method for selecting high-quality pretraining data by leveraging the activation patterns of attention heads in pretrained LLMs. Unlike traditional classifier-based approaches, our method exploits intrinsic model signals to identify reasoning-intensive data, requiring no additional supervision or manual curation. Experimental results on SmolLM-Corpus demonstrate that AttentionInfluence consistently improves downstream performance, selects longer and more diverse high-quality data, and aligns well with existing classifier-based patterns—while offering better weak-to-strong generalization. Our findings suggest that internal model mechanisms can serve as reliable indicators of data quality, offering a scalable and efficient pathway for LLM pretraining.

\clearpage

\bibliographystyle{plainnat}
\bibliography{main}

\begin{thebibliography}{55}
\providecommand{\natexlab}[1]{#1}
\providecommand{\url}[1]{\texttt{#1}}
\expandafter\ifx\csname urlstyle\endcsname\relax
  \providecommand{\doi}[1]{doi: #1}\else
  \providecommand{\doi}{doi: \begingroup \urlstyle{rm}\Url}\fi

\bibitem[Allal et~al.(2025)Allal, Lozhkov, Bakouch, Bl{\'a}zquez, Penedo, Tunstall, Marafioti, Kydl{\'\i}{\v{c}}ek, Lajar{\'\i}n, Srivastav, et~al.]{allal2025smollm2}
Loubna~Ben Allal, Anton Lozhkov, Elie Bakouch, Gabriel~Mart{\'\i}n Bl{\'a}zquez, Guilherme Penedo, Lewis Tunstall, Andr{\'e}s Marafioti, Hynek Kydl{\'\i}{\v{c}}ek, Agust{\'\i}n~Piqueres Lajar{\'\i}n, Vaibhav Srivastav, et~al.
\newblock Smollm2: When smol goes big--data-centric training of a small language model.
\newblock \emph{arXiv preprint arXiv:2502.02737}, 2025.

\bibitem[Ankner et~al.(2024)Ankner, Blakeney, Sreenivasan, Marion, Leavitt, and Paul]{ankner2024perplexed}
Zachary Ankner, Cody Blakeney, Kartik Sreenivasan, Max Marion, Matthew~L Leavitt, and Mansheej Paul.
\newblock Perplexed by perplexity: Perplexity-based data pruning with small reference models.
\newblock \emph{arXiv preprint arXiv:2405.20541}, 2024.

\bibitem[Ben~Allal et~al.(2024)Ben~Allal, Lozhkov, Penedo, Wolf, and von Werra]{benallal2024smollmcorpus}
Loubna Ben~Allal, Anton Lozhkov, Guilherme Penedo, Thomas Wolf, and Leandro von Werra.
\newblock Smollm-corpus, 2024.
\newblock URL \url{https://huggingface.co/datasets/HuggingFaceTB/smollm-corpus}.

\bibitem[Bisk et~al.(2020)Bisk, Zellers, Gao, Choi, et~al.]{bisk2020piqa}
Yonatan Bisk, Rowan Zellers, Jianfeng Gao, Yejin Choi, et~al.
\newblock Piqa: Reasoning about physical commonsense in natural language.
\newblock In \emph{Proceedings of the AAAI conference on artificial intelligence}, volume~34, pages 7432--7439, 2020.

\bibitem[Bricken et~al.(2023)Bricken, Templeton, Batson, Chen, Jermyn, Conerly, Turner, Anil, Denison, Askell, Lasenby, Wu, and et~al.]{anthropic_sparse_autoencoder}
Trenton Bricken, Adly Templeton, Joshua Batson, Brian Chen, Adam Jermyn, Tom Conerly, Nicholas~L Turner, Cem Anil, Carson Denison, Amanda Askell, Robert Lasenby, Yifan Wu, and et~al.
\newblock Towards monosemanticity: Decomposing language models with dictionary learning.
\newblock \url{https://transformer-circuits.pub/2023/monosemantic-features/index.html}, 2023.
\newblock Accessed: 2023-10-04.

\bibitem[Casanueva et~al.(2020)Casanueva, Tem{\v{c}}inas, Gerz, Henderson, and Vuli{\'c}]{casanueva2020efficient}
I{\~n}igo Casanueva, Tadas Tem{\v{c}}inas, Daniela Gerz, Matthew Henderson, and Ivan Vuli{\'c}.
\newblock Efficient intent detection with dual sentence encoders.
\newblock \emph{arXiv preprint arXiv:2003.04807}, 2020.

\bibitem[Chen et~al.(2021)Chen, Tworek, Jun, Yuan, Pinto, Kaplan, Edwards, Burda, Joseph, Brockman, et~al.]{chen2021evaluating}
Mark Chen, Jerry Tworek, Heewoo Jun, Qiming Yuan, Henrique Ponde de~Oliveira Pinto, Jared Kaplan, Harri Edwards, Yuri Burda, Nicholas Joseph, Greg Brockman, et~al.
\newblock Evaluating large language models trained on code.
\newblock \emph{arXiv preprint arXiv:2107.03374}, 2021.

\bibitem[Clark et~al.(2018)Clark, Cowhey, Etzioni, Khot, Sabharwal, Schoenick, and Tafjord]{clark2018think}
Peter Clark, Isaac Cowhey, Oren Etzioni, Tushar Khot, Ashish Sabharwal, Carissa Schoenick, and Oyvind Tafjord.
\newblock Think you have solved question answering? try arc, the ai2 reasoning challenge.
\newblock \emph{arXiv preprint arXiv:1803.05457}, 2018.

\bibitem[Cobbe et~al.(2021)Cobbe, Kosaraju, Bavarian, Chen, Jun, Kaiser, Plappert, Tworek, Hilton, Nakano, et~al.]{cobbe2021training}
Karl Cobbe, Vineet Kosaraju, Mohammad Bavarian, Mark Chen, Heewoo Jun, Lukasz Kaiser, Matthias Plappert, Jerry Tworek, Jacob Hilton, Reiichiro Nakano, et~al.
\newblock Training verifiers to solve math word problems.
\newblock \emph{arXiv preprint arXiv:2110.14168}, 2021.

\bibitem[Dua et~al.(2019)Dua, Wang, Dasigi, Stanovsky, Singh, and Gardner]{dua2019drop}
Dheeru Dua, Yizhong Wang, Pradeep Dasigi, Gabriel Stanovsky, Sameer Singh, and Matt Gardner.
\newblock Drop: A reading comprehension benchmark requiring discrete reasoning over paragraphs.
\newblock \emph{arXiv preprint arXiv:1903.00161}, 2019.

\bibitem[Fu et~al.(2024)Fu, Cai, Asi, Xiong, Dong, and Xiao]{fu2024not}
Yu~Fu, Zefan Cai, Abedelkadir Asi, Wayne Xiong, Yue Dong, and Wen Xiao.
\newblock Not all heads matter: A head-level kv cache compression method with integrated retrieval and reasoning.
\newblock \emph{arXiv preprint arXiv:2410.19258}, 2024.

\bibitem[Geva et~al.(2020)Geva, Schuster, Berant, and Levy]{geva2020transformer}
Mor Geva, Roei Schuster, Jonathan Berant, and Omer Levy.
\newblock Transformer feed-forward layers are key-value memories.
\newblock \emph{arXiv preprint arXiv:2012.14913}, 2020.

\bibitem[Grattafiori et~al.(2024)Grattafiori, Dubey, Jauhri, Pandey, Kadian, Al-Dahle, Letman, Mathur, Schelten, Vaughan, et~al.]{grattafiori2024llama}
Aaron Grattafiori, Abhimanyu Dubey, Abhinav Jauhri, Abhinav Pandey, Abhishek Kadian, Ahmad Al-Dahle, Aiesha Letman, Akhil Mathur, Alan Schelten, Alex Vaughan, et~al.
\newblock The llama 3 herd of models.
\newblock \emph{arXiv preprint arXiv:2407.21783}, 2024.

\bibitem[Held et~al.(2025)Held, Paranjape, Koura, Lewis, Zhang, and Mihaylov]{held2025optimizing}
William Held, Bhargavi Paranjape, Punit~Singh Koura, Mike Lewis, Frank Zhang, and Todor Mihaylov.
\newblock Optimizing pretraining data mixtures with llm-estimated utility.
\newblock \emph{arXiv preprint arXiv:2501.11747}, 2025.

\bibitem[Hendrycks et~al.(2020)Hendrycks, Burns, Basart, Zou, Mazeika, Song, and Steinhardt]{hendrycks2020measuring}
Dan Hendrycks, Collin Burns, Steven Basart, Andy Zou, Mantas Mazeika, Dawn Song, and Jacob Steinhardt.
\newblock Measuring massive multitask language understanding.
\newblock \emph{arXiv preprint arXiv:2009.03300}, 2020.

\bibitem[Hendrycks et~al.(2021)Hendrycks, Burns, Kadavath, Arora, Basart, Tang, Song, and Steinhardt]{hendrycks2021measuring}
Dan Hendrycks, Collin Burns, Saurav Kadavath, Akul Arora, Steven Basart, Eric Tang, Dawn Song, and Jacob Steinhardt.
\newblock Measuring mathematical problem solving with the math dataset.
\newblock \emph{arXiv preprint arXiv:2103.03874}, 2021.

\bibitem[Hinton et~al.(2015)Hinton, Vinyals, and Dean]{hinton2015distilling}
Geoffrey Hinton, Oriol Vinyals, and Jeff Dean.
\newblock Distilling the knowledge in a neural network.
\newblock \emph{arXiv preprint arXiv:1503.02531}, 2015.

\bibitem[Hu et~al.(2024)Hu, Tu, Han, He, Cui, Long, Zheng, Fang, Huang, Zhao, et~al.]{hu2024minicpm}
Shengding Hu, Yuge Tu, Xu~Han, Chaoqun He, Ganqu Cui, Xiang Long, Zhi Zheng, Yewei Fang, Yuxiang Huang, Weilin Zhao, et~al.
\newblock Minicpm: Unveiling the potential of small language models with scalable training strategies.
\newblock \emph{arXiv preprint arXiv:2404.06395}, 2024.

\bibitem[Huang et~al.(2023)Huang, Bai, Zhu, Zhang, Zhang, Su, Liu, Lv, Zhang, Fu, et~al.]{huang2023c}
Yuzhen Huang, Yuzhuo Bai, Zhihao Zhu, Junlei Zhang, Jinghan Zhang, Tangjun Su, Junteng Liu, Chuancheng Lv, Yikai Zhang, Yao Fu, et~al.
\newblock C-eval: A multi-level multi-discipline chinese evaluation suite for foundation models.
\newblock \emph{Advances in Neural Information Processing Systems}, 36:\penalty0 62991--63010, 2023.

\bibitem[Joshi et~al.(2017)Joshi, Choi, Weld, and Zettlemoyer]{joshi2017triviaqa}
Mandar Joshi, Eunsol Choi, Daniel~S Weld, and Luke Zettlemoyer.
\newblock Triviaqa: A large scale distantly supervised challenge dataset for reading comprehension.
\newblock \emph{arXiv preprint arXiv:1705.03551}, 2017.

\bibitem[Joulin et~al.(2016)Joulin, Grave, Bojanowski, Douze, J{\'e}gou, and Mikolov]{joulin2016fasttext}
Armand Joulin, Edouard Grave, Piotr Bojanowski, Matthijs Douze, H{\'e}rve J{\'e}gou, and Tomas Mikolov.
\newblock Fasttext. zip: Compressing text classification models.
\newblock \emph{arXiv preprint arXiv:1612.03651}, 2016.

\bibitem[Ko et~al.(2024)Ko, Kang, Shi, Jin, Yu, and Jia]{ko2024mirrored}
Myeongseob Ko, Feiyang Kang, Weiyan Shi, Ming Jin, Zhou Yu, and Ruoxi Jia.
\newblock The mirrored influence hypothesis: Efficient data influence estimation by harnessing forward passes.
\newblock In \emph{Proceedings of the IEEE/CVF Conference on Computer Vision and Pattern Recognition}, pages 26286--26295, 2024.

\bibitem[Lai et~al.(2017)Lai, Xie, Liu, Yang, and Hovy]{lai2017racelargescalereadingcomprehension}
Guokun Lai, Qizhe Xie, Hanxiao Liu, Yiming Yang, and Eduard Hovy.
\newblock Race: Large-scale reading comprehension dataset from examinations, 2017.
\newblock URL \url{https://arxiv.org/abs/1704.04683}.

\bibitem[Li et~al.(2024{\natexlab{a}})Li, Fang, Smyrnis, Ivgi, Jordan, Gadre, Bansal, Guha, Keh, Arora, et~al.]{li2024datacomp}
Jeffrey Li, Alex Fang, Georgios Smyrnis, Maor Ivgi, Matt Jordan, Samir~Yitzhak Gadre, Hritik Bansal, Etash Guha, Sedrick~Scott Keh, Kushal Arora, et~al.
\newblock Datacomp-lm: In search of the next generation of training sets for language models.
\newblock \emph{Advances in Neural Information Processing Systems}, 37:\penalty0 14200--14282, 2024{\natexlab{a}}.

\bibitem[Li et~al.(2024{\natexlab{b}})Li, Wei, Zhang, Yu, Hu, and Peng]{li2024scalingfilter}
Ruihang Li, Yixuan Wei, Miaosen Zhang, Nenghai Yu, Han Hu, and Houwen Peng.
\newblock Scalingfilter: Assessing data quality through inverse utilization of scaling laws.
\newblock \emph{arXiv preprint arXiv:2408.08310}, 2024{\natexlab{b}}.

\bibitem[Lin et~al.(2024)Lin, Gou, Gong, Liu, Shen, Xu, Lin, Yang, Jiao, Duan, et~al.]{lin2024rho}
Zhenghao Lin, Zhibin Gou, Yeyun Gong, Xiao Liu, Yelong Shen, Ruochen Xu, Chen Lin, Yujiu Yang, Jian Jiao, Nan Duan, et~al.
\newblock Rho-1: Not all tokens are what you need.
\newblock \emph{arXiv preprint arXiv:2404.07965}, 2024.

\bibitem[Liu et~al.(2024)Liu, Zheng, Muennighoff, Zeng, Dou, Pang, Jiang, and Lin]{liu2024regmix}
Qian Liu, Xiaosen Zheng, Niklas Muennighoff, Guangtao Zeng, Longxu Dou, Tianyu Pang, Jing Jiang, and Min Lin.
\newblock Regmix: Data mixture as regression for language model pre-training.
\newblock \emph{arXiv preprint arXiv:2407.01492}, 2024.

\bibitem[Lv et~al.(2024)Lv, Chen, Zhang, Wang, Liu, Wen, Xie, and Yan]{lv2024interpreting}
Ang Lv, Yuhan Chen, Kaiyi Zhang, Yulong Wang, Lifeng Liu, Ji-Rong Wen, Jian Xie, and Rui Yan.
\newblock Interpreting key mechanisms of factual recall in transformer-based language models.
\newblock \emph{arXiv preprint arXiv:2403.19521}, 2024.

\bibitem[Mihaylov et~al.(2018)Mihaylov, Clark, Khot, and Sabharwal]{mihaylov2018can}
Todor Mihaylov, Peter Clark, Tushar Khot, and Ashish Sabharwal.
\newblock Can a suit of armor conduct electricity? a new dataset for open book question answering.
\newblock \emph{arXiv preprint arXiv:1809.02789}, 2018.

\bibitem[OLMo et~al.(2024)OLMo, Walsh, Soldaini, Groeneveld, Lo, Arora, Bhagia, Gu, Huang, Jordan, et~al.]{olmo20242}
Team OLMo, Pete Walsh, Luca Soldaini, Dirk Groeneveld, Kyle Lo, Shane Arora, Akshita Bhagia, Yuling Gu, Shengyi Huang, Matt Jordan, et~al.
\newblock 2 olmo 2 furious.
\newblock \emph{arXiv preprint arXiv:2501.00656}, 2024.

\bibitem[Olsson et~al.(2022)Olsson, Elhage, Nanda, Joseph, DasSarma, Henighan, Mann, Askell, Bai, Chen, et~al.]{olsson2022context}
Catherine Olsson, Nelson Elhage, Neel Nanda, Nicholas Joseph, Nova DasSarma, Tom Henighan, Ben Mann, Amanda Askell, Yuntao Bai, Anna Chen, et~al.
\newblock In-context learning and induction heads.
\newblock \emph{arXiv preprint arXiv:2209.11895}, 2022.

\bibitem[Penedo et~al.(2024)Penedo, Kydl{\'\i}{\v{c}}ek, Lozhkov, Mitchell, Raffel, Von~Werra, Wolf, et~al.]{penedo2024fineweb}
Guilherme Penedo, Hynek Kydl{\'\i}{\v{c}}ek, Anton Lozhkov, Margaret Mitchell, Colin~A Raffel, Leandro Von~Werra, Thomas Wolf, et~al.
\newblock The fineweb datasets: Decanting the web for the finest text data at scale.
\newblock \emph{Advances in Neural Information Processing Systems}, 37:\penalty0 30811--30849, 2024.

\bibitem[Peng et~al.(2025)Peng, Yang, Zeng, Lin, Liu, and Zhao]{peng2025dataman}
Ru~Peng, Kexin Yang, Yawen Zeng, Junyang Lin, Dayiheng Liu, and Junbo Zhao.
\newblock Dataman: Data manager for pre-training large language models.
\newblock \emph{arXiv preprint arXiv:2502.19363}, 2025.

\bibitem[Qiu et~al.(2024)Qiu, Li, Huang, Jiao, Zhong, and King]{qiu2024clongeval}
Zexuan Qiu, Jingjing Li, Shijue Huang, Xiaoqi Jiao, Wanjun Zhong, and Irwin King.
\newblock Clongeval: A chinese benchmark for evaluating long-context large language models.
\newblock \emph{arXiv preprint arXiv:2403.03514}, 2024.

\bibitem[Rae et~al.(2021)Rae, Borgeaud, Cai, Millican, Hoffmann, Song, Aslanides, Henderson, Ring, Young, et~al.]{rae2021scaling}
Jack~W Rae, Sebastian Borgeaud, Trevor Cai, Katie Millican, Jordan Hoffmann, Francis Song, John Aslanides, Sarah Henderson, Roman Ring, Susannah Young, et~al.
\newblock Scaling language models: Methods, analysis \& insights from training gopher.
\newblock \emph{arXiv preprint arXiv:2112.11446}, 2021.

\bibitem[Reimers and Gurevych(2019)]{reimers2019sentence}
Nils Reimers and Iryna Gurevych.
\newblock Sentence-bert: Sentence embeddings using siamese bert-networks.
\newblock \emph{arXiv preprint arXiv:1908.10084}, 2019.

\bibitem[Rein et~al.(2023)Rein, Hou, Stickland, Petty, Pang, Dirani, Michael, and Bowman]{rein2023gpqagraduatelevelgoogleproofqa}
David Rein, Betty~Li Hou, Asa~Cooper Stickland, Jackson Petty, Richard~Yuanzhe Pang, Julien Dirani, Julian Michael, and Samuel~R. Bowman.
\newblock Gpqa: A graduate-level google-proof q\&a benchmark, 2023.
\newblock URL \url{https://arxiv.org/abs/2311.12022}.

\bibitem[Ruis et~al.(2024)Ruis, Mozes, Bae, Kamalakara, Talupuru, Locatelli, Kirk, Rockt{\"a}schel, Grefenstette, and Bartolo]{ruis2024procedural}
Laura Ruis, Maximilian Mozes, Juhan Bae, Siddhartha~Rao Kamalakara, Dwarak Talupuru, Acyr Locatelli, Robert Kirk, Tim Rockt{\"a}schel, Edward Grefenstette, and Max Bartolo.
\newblock Procedural knowledge in pretraining drives reasoning in large language models.
\newblock \emph{arXiv preprint arXiv:2411.12580}, 2024.

\bibitem[Sakaguchi et~al.(2021)Sakaguchi, Bras, Bhagavatula, and Choi]{sakaguchi2021winogrande}
Keisuke Sakaguchi, Ronan~Le Bras, Chandra Bhagavatula, and Yejin Choi.
\newblock Winogrande: An adversarial winograd schema challenge at scale.
\newblock \emph{Communications of the ACM}, 64\penalty0 (9):\penalty0 99--106, 2021.

\bibitem[Su et~al.(2024)Su, Kong, Lin, Jennings, Norick, Kliegl, Patwary, Shoeybi, and Catanzaro]{su2024nemotron}
Dan Su, Kezhi Kong, Ying Lin, Joseph Jennings, Brandon Norick, Markus Kliegl, Mostofa Patwary, Mohammad Shoeybi, and Bryan Catanzaro.
\newblock Nemotron-cc: Transforming common crawl into a refined long-horizon pretraining dataset.
\newblock \emph{arXiv preprint arXiv:2412.02595}, 2024.

\bibitem[Suzgun et~al.(2022)Suzgun, Scales, Sch{\"a}rli, Gehrmann, Tay, Chung, Chowdhery, Le, Chi, Zhou, et~al.]{suzgun2022challenging}
Mirac Suzgun, Nathan Scales, Nathanael Sch{\"a}rli, Sebastian Gehrmann, Yi~Tay, Hyung~Won Chung, Aakanksha Chowdhery, Quoc~V Le, Ed~H Chi, Denny Zhou, et~al.
\newblock Challenging big-bench tasks and whether chain-of-thought can solve them.
\newblock \emph{arXiv preprint arXiv:2210.09261}, 2022.

\bibitem[Talmor et~al.(2018)Talmor, Herzig, Lourie, and Berant]{talmor2018commonsenseqa}
Alon Talmor, Jonathan Herzig, Nicholas Lourie, and Jonathan Berant.
\newblock Commonsenseqa: A question answering challenge targeting commonsense knowledge.
\newblock \emph{arXiv preprint arXiv:1811.00937}, 2018.

\bibitem[Touvron et~al.(2023)Touvron, Lavril, Izacard, Martinet, Lachaux, Lacroix, Rozi{\`e}re, Goyal, Hambro, Azhar, et~al.]{touvron2023llama}
Hugo Touvron, Thibaut Lavril, Gautier Izacard, Xavier Martinet, Marie-Anne Lachaux, Timoth{\'e}e Lacroix, Baptiste Rozi{\`e}re, Naman Goyal, Eric Hambro, Faisal Azhar, et~al.
\newblock Llama: Open and efficient foundation language models.
\newblock \emph{arXiv preprint arXiv:2302.13971}, 2023.

\bibitem[Wang et~al.(2024)Wang, Ma, Zhang, Ni, Chandra, Guo, Ren, Arulraj, He, Jiang, et~al.]{wang2024mmlu}
Yubo Wang, Xueguang Ma, Ge~Zhang, Yuansheng Ni, Abhranil Chandra, Shiguang Guo, Weiming Ren, Aaran Arulraj, Xuan He, Ziyan Jiang, et~al.
\newblock Mmlu-pro: A more robust and challenging multi-task language understanding benchmark.
\newblock In \emph{The Thirty-eight Conference on Neural Information Processing Systems Datasets and Benchmarks Track}, 2024.

\bibitem[Wettig et~al.(2024)Wettig, Gupta, Malik, and Chen]{wettig2024qurating}
Alexander Wettig, Aatmik Gupta, Saumya Malik, and Danqi Chen.
\newblock Qurating: Selecting high-quality data for training language models.
\newblock \emph{arXiv preprint arXiv:2402.09739}, 2024.

\bibitem[Wettig et~al.(2025)Wettig, Lo, Min, Hajishirzi, Chen, and Soldaini]{wettig2025organize}
Alexander Wettig, Kyle Lo, Sewon Min, Hannaneh Hajishirzi, Danqi Chen, and Luca Soldaini.
\newblock Organize the web: Constructing domains enhances pre-training data curation.
\newblock \emph{arXiv preprint arXiv:2502.10341}, 2025.

\bibitem[Wu et~al.(2024)Wu, Wang, Xiao, Peng, and Fu]{wu2024retrieval}
Wenhao Wu, Yizhong Wang, Guangxuan Xiao, Hao Peng, and Yao Fu.
\newblock Retrieval head mechanistically explains long-context factuality.
\newblock \emph{arXiv preprint arXiv:2404.15574}, 2024.

\bibitem[Xie et~al.(2023)Xie, Pham, Dong, Du, Liu, Lu, Liang, Le, Ma, and Yu]{xie2023doremi}
Sang~Michael Xie, Hieu Pham, Xuanyi Dong, Nan Du, Hanxiao Liu, Yifeng Lu, Percy~S Liang, Quoc~V Le, Tengyu Ma, and Adams~Wei Yu.
\newblock Doremi: Optimizing data mixtures speeds up language model pretraining.
\newblock \emph{Advances in Neural Information Processing Systems}, 36:\penalty0 69798--69818, 2023.

\bibitem[Ye et~al.(2024)Ye, Liu, Sun, Zhan, Zhou, and Qiu]{ye2024data}
Jiasheng Ye, Peiju Liu, Tianxiang Sun, Jun Zhan, Yunhua Zhou, and Xipeng Qiu.
\newblock Data mixing laws: Optimizing data mixtures by predicting language modeling performance.
\newblock \emph{arXiv preprint arXiv:2403.16952}, 2024.

\bibitem[Yu et~al.(2024)Yu, Das, and Xiong]{yu2024mates}
Zichun Yu, Spandan Das, and Chenyan Xiong.
\newblock Mates: Model-aware data selection for efficient pretraining with data influence models.
\newblock \emph{Advances in Neural Information Processing Systems}, 37:\penalty0 108735--108759, 2024.

\bibitem[Zellers et~al.(2019)Zellers, Holtzman, Bisk, Farhadi, and Choi]{zellers2019hellaswag}
Rowan Zellers, Ari Holtzman, Yonatan Bisk, Ali Farhadi, and Yejin Choi.
\newblock Hellaswag: Can a machine really finish your sentence?
\newblock \emph{arXiv preprint arXiv:1905.07830}, 2019.

\bibitem[Zhao et~al.(2024)Zhao, Thai, Zhang, Hu, Ba, Zhou, Cai, Liu, and Sun]{zhao2024decoratelm}
Ranchi Zhao, Zhen~Leng Thai, Yifan Zhang, Shengding Hu, Yunqi Ba, Jie Zhou, Jie Cai, Zhiyuan Liu, and Maosong Sun.
\newblock Decoratelm: Data engineering through corpus rating, tagging, and editing with language models.
\newblock \emph{arXiv preprint arXiv:2410.05639}, 2024.

\bibitem[Zheng et~al.(2024)Zheng, Wang, Huang, Song, Yang, Tang, Xiong, and Li]{zheng2024attention}
Zifan Zheng, Yezhaohui Wang, Yuxin Huang, Shichao Song, Mingchuan Yang, Bo~Tang, Feiyu Xiong, and Zhiyu Li.
\newblock Attention heads of large language models: A survey.
\newblock \emph{arXiv preprint arXiv:2409.03752}, 2024.

\bibitem[Zhong et~al.(2023)Zhong, Cui, Guo, Liang, Lu, Wang, Saied, Chen, and Duan]{zhong2023agieval}
Wanjun Zhong, Ruixiang Cui, Yiduo Guo, Yaobo Liang, Shuai Lu, Yanlin Wang, Amin Saied, Weizhu Chen, and Nan Duan.
\newblock Agieval: A human-centric benchmark for evaluating foundation models.
\newblock \emph{arXiv preprint arXiv:2304.06364}, 2023.

\bibitem[Zhu et~al.(2025)Zhu, Li, Wang, Haehn, and Liang]{zhu2025focus}
Youxiang Zhu, Ruochen Li, Danqing Wang, Daniel Haehn, and Xiaohui Liang.
\newblock Focus directions make your language models pay more attention to relevant contexts.
\newblock \emph{arXiv preprint arXiv:2503.23306}, 2025.

\end{thebibliography}

\clearpage

\beginappendix

\newtcolorbox[auto counter, number within=section]{methodbox}[2][]{
  colback=white, 
  colframe=teal!80!green!80!black,  
  width=\textwidth,
  arc=2mm, 
  boxrule=0.5mm, 
  title={\normalsize\faWrench\hspace{0.5em}#2}, 
  breakable,
  fonttitle=\bfseries\Large, 
  fontupper=\small, 
  #1
}

\newtcolorbox[auto counter, number within=section]{methodbox2}[2][]{
  colback=white, 
  colframe=blue!70!cyan!80!black,  
  width=\textwidth,
  arc=2mm, 
  boxrule=0.5mm, 
  title={\normalsize\faWrench\hspace{0.5em}#2}, 
  breakable,
  fonttitle=\bfseries\Large, 
  fontupper=\small, 
  #1
}

\newtcolorbox[auto counter, number within=section]{methodbox3}[2][]{
  colback=white, 
  colframe=violet!80!black,  
  width=\textwidth,
  arc=2mm, 
  boxrule=0.5mm, 
  title={\normalsize\faWrench\hspace{0.5em}#2}, 
  breakable,
  fonttitle=\bfseries\Large, 
  fontupper=\small, 
  #1
}

\newtcolorbox[auto counter, number within=section]{methodbox4}[2][]{
  colback=white, 
  colframe=red!70!black,  
  width=\textwidth,
  arc=2mm, 
  boxrule=0.5mm, 
  title={\normalsize\faWrench\hspace{0.5em}#2}, 
  breakable,
  fonttitle=\bfseries\Large, 
  fontupper=\small, 
  #1
}

\newtcolorbox[auto counter, number within=section]{methodbox5}[2][]{
  colback=white, 
  colframe=yellow!70!black,  
  width=\textwidth,
  arc=2mm, 
  boxrule=0.5mm, 
  title={\normalsize\faWrench\hspace{0.5em}#2}, 
  breakable,
  fonttitle=\bfseries\Large, 
  fontupper=\small, 
  #1
}

\section{Synthetic Test Sample}
\label{sec:synthetic_test_sample}
\begin{framed}
\begin{verbatim}
model input:  
Please extract the value corresponding to the specified key from the 
following JSON object. Output only the value of the corresponding key 
and nothing else. The JSON data is as follows:  
{context}

{question-shot1}  
{answer-shot1}  
{question-shot2}  
{answer-shot2}  
{question-shot3}  
{answer-shot3}  
{question}

answer:  
{answer}
\end{verbatim}
\end{framed}

\section{Evolution of Retrieval Heads in Pretrained Models}
We apply the method described in Section \ref{sec:detect} to identify retrieval heads at six checkpoints of the pretrained 1.3B-parameter model. These checkpoints correspond to training progress at 5B, 307B, 608B, 898B, 1200B, and 1499B tokens, respectively.

\section{Masking Operation}
\label{sec:Masking_Operation}
The "mask" operation is to set the attention weights provided by the specific attention heads to equal weights. And if the length of the sequence is $L$, the attention weight of each token should be set to $\frac{1}{L}$. 


\section{Effect of Masking Retrieval Heads vs. Random Non-Retrieval Heads}
\label{appd:different_heads_influence}
\begin{table*}[h]
    \centering
    \resizebox{0.8\textwidth}{!}{
    \begin{tabular}{*{7}{c}}
        \toprule
        Model & \multicolumn{6}{c}{Benchmarks} \\
        
        \midrule\midrule
        \multirow{4}{*}{1.3B}  &   Hellaswag &WinoGrande  &MMLU &MMLU-Pro & AGIEval-en & GPQA \\
        & 0.5715
  &0.6062 & 0.4258 &  0.1290 & 0.2047 & 0.2203\\
        & DROP &BBH & GSM8K & HumanEval & Banking77-en-ICL \\
        & 0.2344 & 0.3166 & 0.1820 &0.1707 & 0.4148 & \\ \midrule\midrule

        \multirow{4}{*}{1.3B (Random Masked, Non-Retrieval Heads)}  &   Hellaswag &WinoGrande  &MMLU &MMLU-Pro & AGIEval-en & GPQA \\
        & 0.5518
 &0.6069 & 0.4165 &  0.1275 & 0.2072 & 0.2071\\
        & DROP &BBH & GSM8K & HumanEval & Banking77-en-ICL \\
        & 0.2190 & 0.3005 & 0.1274  &0.1159 & 0.3840 & \\ \midrule\midrule
        \multirow{4}{*}{1.3B (Masked, Retrieval Heads)}  &   Hellaswag &WinoGrande  &MMLU &MMLU-Pro & AGIEval-en & GPQA \\
        & 0.5493 &0.5801 &0.3089 &  0.0305 & 0.1298 & 0.1827\\
        & DROP &BBH & GSM8K & HumanEval & Banking77-en-ICL \\
        & 0.1141 & 0.0429 & 0.0068 &0.1098 &0.0001 & \\

    \bottomrule
    
    \end{tabular}
    }
    \caption{Effect of Masking Retrieval Heads vs. Random Non-Retrieval Heads on Reasoning and In-Context Learning}
  
    \label{tab:effect_of_heads}
\end{table*}
Banking77-en-ICL is an internal evaluation task for assessing a model's in-context learning ability. It requires models to perform many-shot classification on the Banking77-en dataset\cite{casanueva2020efficient}. Here, "Masked, Retrieval Heads" refers to masking attention heads ranked in the top 5\% by retrieval score, while "Random Masked, Non-Retrieval Heads" refers to randomly masking heads ranked between the top 5\% and top 100\% (i.e., the remaining 95\%) by retrieval score.
We conduct the experiments using the models shown in the \autoref{tab:hyperparams_models_used_by_attention_influence} and find that masking retrieval heads significantly impairs the model’s reasoning performance, while masking random non-retrieval heads has only a minor effect---consistent with the findings of \citet{wu2024retrieval}. In addition, we find that retrieval heads also play an essential role in the model's in-context learning ability.

\section{Experiment Setting}
\label{Sec: Experiment Setting}

\paragraph{Pretraining Data}
To ensure reproducibility, we use SmolLM-Corpus\cite{benallal2024smollmcorpus} as the pretraining dataset. The composition of the SmolLM-Corpus dataset is shown in the \autoref{tab:smollm_corpus}. We sample 100 million tokens from SmolLM-Corpus as the validation dataset.

\paragraph{Pretrained models used by AttentionInfluence}
\label{sec:models_used_by_attention_influence}
In this work, AttentionInfluence employs internal pretrained models based on the LLaMA2-alike architecture. The hyperparameters of the models are detailed in \autoref{tab:hyperparams_models_used_by_attention_influence}.

\begin{table*}[h]
    \centering
    \setlength{\tabcolsep}{4pt}
    
    \resizebox{0.6\textwidth}{!}{
    \begin{tabular}{*{11}{c}}
        \toprule
        model &pretraining &vocab & hidden & ffn & num & num & shared & seq & tie \\
        size & tokens &size & size & inner & heads & layers & q\_head & len & emb \\
        \midrule
        {1.3B}  & 1.5TB  &155136 & 2,560 &  10,240 & 20 & 16 & 2 & 4,096 & true \\
        {7B}  & 9TB &155136 & 4,096 &  16,384 & 32 & 32 & 2 & 8,192 & true  \\
        \bottomrule
    
    \end{tabular}
    }
    \caption{Hyperparams of the Pretrained Models Used by AttentionInfluence.}
    \label{tab:hyperparams_models_used_by_attention_influence}
\end{table*}

\paragraph{Model trained in the experiment}
\label{sec:appd_training}
The hyperparameters are presented in \autoref{tab:model_hyperparams}, and tokenizer used for training and computing token counts is the same as 
SmolLM\footnote{\url{https://huggingface.co/HuggingFaceTB/cosmo2-tokenizer}} with a vocab size of 49,152.

\paragraph{Pretraining setting} Referring to SmolLM\cite{benallal2024smollmcorpus}, our experiments are conducted with WSD learning rate scheduler~\citep{hu2024minicpm} with 0.1\% warmup steps, 75\% stable phase, and a final 25\% decay phase. The amount of training tokens is 1 TB. The training is distributed across 32 machines, each equipped with eight H100-80GB GPUs.
\begin{table*}[ht]
  \centering
  \resizebox{0.7\textwidth}{!}{
  \begin{tabular}{lccccc}
    \toprule
    {Dataset} & {FineWeb-Edu-Dedup} & {Cosmopedia-V2} & {Python-Edu} & {OpenWebMath} \\
    \midrule
    \# Tokens (billions) & 193.3 & 27.9 & 3.8 & 13.3 \\
    \bottomrule
  \end{tabular}
  }
  \caption{\label{tab:smollm_corpus}
    Composition of the SmolLM Corpus Dataset.
  }
\end{table*}

\begin{table*}[h]
    \centering
    \resizebox{0.75\textwidth}{!}{
    \begin{tabular}{*{11}{c}}
        \toprule
        model & batch & learning & hidden & ffn & num & num & shared & seq & tie & total \\
        size & size & rate & size & inner & heads & layers & q\_head & len & emb & params \\
        \midrule
        {7B}  & 1,024 & 4e-4   & 4,096 &  8,192 & 32 & 32 & 4 & 8,192 & false & 6.98B \\
        \bottomrule
    \end{tabular}
    }
    \caption{Hyperparams of the Model Trained in the Experiment.}
    \label{tab:model_hyperparams}
\end{table*}

\paragraph{Evaluation details}
\label{appd:evaluation_details}
To ensure that all demonstrations, along with the question and the generated prediction, fit within the 8192-token context window, we use a different number of few-shot examples per evaluation task. Specifically, we use the following numbers of demonstrations (in parentheses): MATH (\texttt{minerva\_math}) (4), DROP (3), BBH (3), and 5 for all other tasks. We report accuracy for most tasks, with the following exceptions: \texttt{exact\_match} for MMLU-Pro, TriviaQA, and BBH; \texttt{flexible-extract} for GSM8K; \texttt{math\_verify} for MATH; and F1 score for DROP. When available, we use the normalized accuracy (\texttt{acc\_norm}) metric provided by the lm-evaluation-harness. ARC(C+E) denotes the average accuracy over ARC-Challenge (ARC-C) and ARC-Easy (ARC-E). For specific tasks, we adopt the following exceptions:
\begin{itemize}[leftmargin=1.5em]
\item
For AGIEval, we conduct the official few-shot evaluation using the official AGIEval repository\footnote{https://github.com/ruixiangcui/AGIEval/tree/main}.
\item
For HumanEval, we conduct zero-shot evaluation using the BigCode evaluation harness\footnote{https://github.com/bigcode-project/bigcode-evaluation-harness} and report pass@1 using the following generation settings, which are the same as those used in SmolLM\cite{benallal2024smollmcorpus}: temperature = 0.2, top-p = 0.95, \texttt{n\_samples} = 20, and \texttt{max\_length\_generation} = 1024.
\item
For DROP, we fix a known bug in the lm-evaluation-harness implementation, following the discussion\footnote{https://github.com/EleutherAI/lm-evaluation-harness/issues/2137}.
\end{itemize}

\section{Detailed Performance Evolution During Pretraining}
\label{sec:extra_exp}
\begin{figure*}[!tb]
    \centering
    \includegraphics[width=0.55\linewidth]{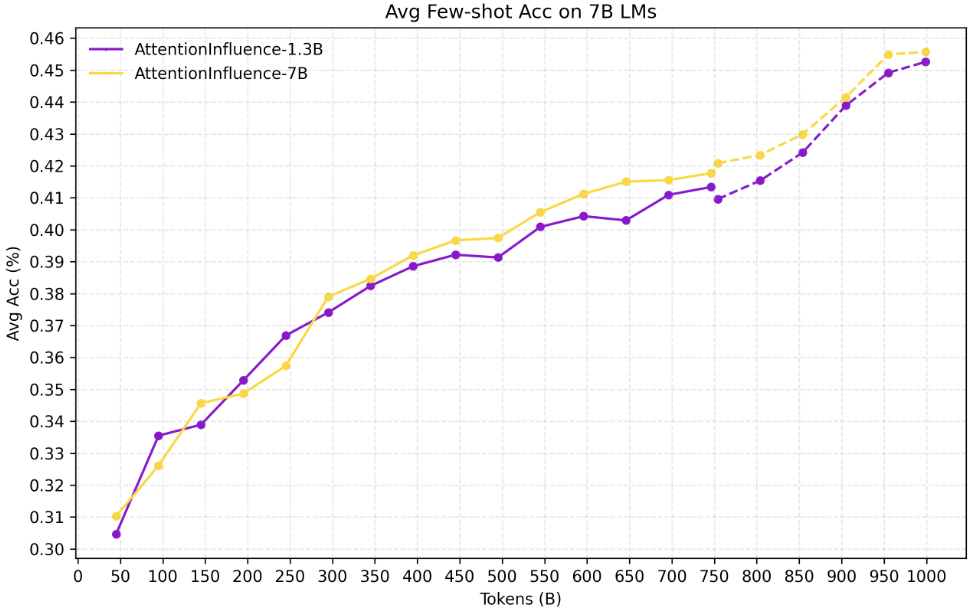}
\caption{Performance evolution on comprehensive benchmark evaluations during pretraining. The first 750 billion tokens correspond to the pretraining phase, represented by solid lines, while the subsequent 250 billion tokens represent the learning rate annealing phase, represented by dashed lines, using the same dataset.
After around 100 billion tokens, {AttentionInfluence-1.3B} consistently outperforms the baseline across a wide range of tasks on average, including the annealing phase.}
    \label{fig:performance_evolution_scaling}
\end{figure*}

\begin{figure*}[!tb]
    \centering
\includegraphics[width=0.62\linewidth]{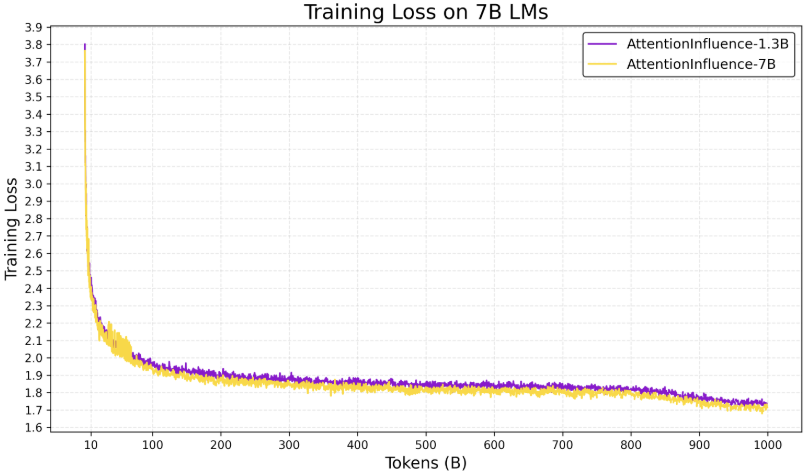}
\caption{
Training loss}
    \label{fig:training_loss_2}
\end{figure*}

\begin{figure*}[!tb]
    \centering
    \includegraphics[width=0.6\linewidth]{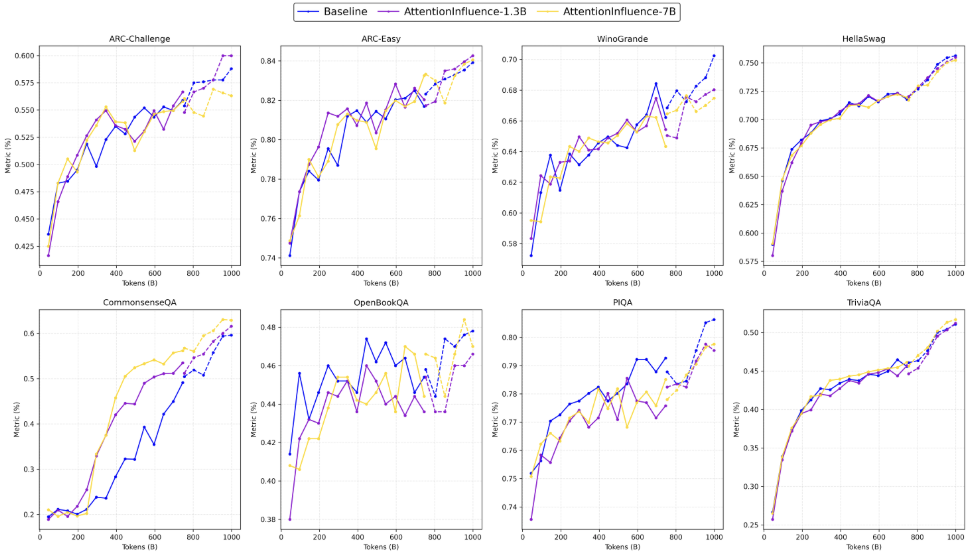}
\caption{
The performance evolution during pretraining on relatively simple benchmarks (i.e., ARC-Challenge, ARC-Easy, WinoGrande, HellaSwag, CommonsenseQA, OpenBookQA, PIQA, TirvialQA).
The first 750 billion tokens correspond to the standard pretraining phase (solid lines), followed by 250 billion tokens under learning rate annealing (dashed lines).
Curves with the same color (solid and dashed) indicate training on the same dataset.
After approximately 100 billion tokens, AttentionInfluence-1.3B consistently outperforms the baseline across a broad range of tasks, including during the annealing phase.
}
    \label{fig:performance_evolution_full_1}
\end{figure*}

\begin{figure*}[!tb]
    \centering
    \includegraphics[width=0.6\linewidth]{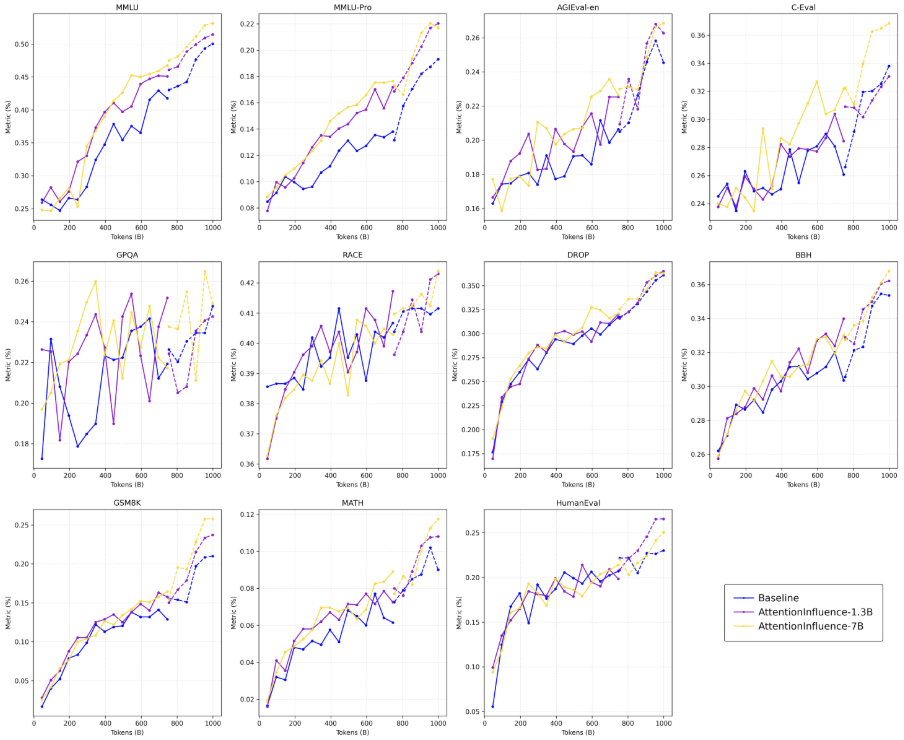}
\caption{
The performance evolution during pretraining on knowledge-intensive and reasoning-heavy benchmarks (i.e., MMLU, MMLU-Pro, AGIEval-en, C-Eval, GPQA, RACE, DROP, BBH, GSM8K, MATH, and HumanEval).
The first 750 billion tokens correspond to the standard pretraining phase (solid lines), followed by 250 billion tokens under learning rate annealing (dashed lines).
Curves with the same color (solid and dashed) indicate training on the same dataset.
After around 100 billion tokens, {AttentionInfluence-1.3B} consistently outperforms the baseline across a wide range of tasks on average, including the annealing phase.}
    \label{fig:performance_evolution_full_2}
\end{figure*}

As shown in \autoref{fig:performance_evolution_scaling}, \autoref{fig:performance_evolution_full_1}, and \autoref{fig:performance_evolution_full_2}, we illustrate how the performance of the baseline, the 1.3B method, and the 7B method evolves across different benchmarks as the number of training tokens increases.
In addition, panel (b) of \autoref{fig:performance_evolution} and \autoref{fig:training_loss_2} present the training loss comparison among baseline, AttentionInfluence-1.3B, and AttentionInfluence-7B.
Furthermore, we report the evaluation results of LLMs trained on data selected by AttentionInfluence-1.3B and AttentionInfluence-7B, as shown in \autoref{tab:abalation_result}.

\begin{table}[htbp]
\centering
\resizebox{0.9\textwidth}{!}{%
\begin{tabular}{llccccccccc}
\toprule
Model  &	\#Tokens & Avg. & \multicolumn{6}{c}{Metrics}
\\
\bottomrule \bottomrule

\multirow{6}{*}{AttentionInfluence-1.3B w/o LRD} & \multirow{6}{*}{746B}  & \multirow{6}{*}{41.34}   &    

ARC-C &	 ARC-E &	 ARC(C+E) &	 Wino. 	& Hella. &	 CSQA & OpenBookQA  \\
 & & & 56.66 &	 82.03 	& 69.35 &	 {65.43}&	 71.90 &	 53.48 & {43.60} \\
 & & & PIQA& 	 TriviaQA&	 MMLU   &  MMLU-Pro &	AGIEval-en & RACE 	 &DROP 	   \\
 & &  & 77.58 &	 45.68	 & 45.10  & 17.19 &	22.50	& 41.72 	& 32.03 \\
&  &  & BBH &	 GSM8K& 	 MATH &	 HumanEval & C-Eval & GPQA  \\
 & &   & 33.99 	& 15.77&	 \ 7.25 	 &19.85 & 28.45& 25.18

 \\
 \midrule\midrule

 \multirow{6}{*}{AttentionInfluence-7B w/o LRD}   	& \multirow{6}{*}{746B}  &   \multirow{6}{*}{\colorbox{green!15}{41.77}}  &   

ARC-C &	 ARC-E &	 ARC(C+E) &	 Wino. 	& Hella. &	 CSQA & OpenBookQA \\
 & & & 55.80 & \colorbox{green!15}{83.25}& \colorbox{green!15}{69.53}& 64.33& \colorbox{green!15}{71.94}& \colorbox{green!15}{56.18}& \colorbox{green!15}{44.40}	 \\
 & & & PIQA& 	 TriviaQA&	 MMLU   &  MMLU-Pro &	AGIEval-en & RACE 	 &DROP 	   \\
 & & &    \colorbox{green!15}{78.51}& \colorbox{green!15}{46.14}& \colorbox{green!15}{46.77}& \colorbox{green!15}{17.64}& \colorbox{green!15}{22.64}& 40.29& \colorbox{green!15}{32.09}   \\
&  & & BBH &	 GSM8K& 	 MATH &	 HumanEval & C-Eval & GPQA \\
& & &      33.02& \colorbox{green!15}{16.45}& 6.78& \colorbox{green!15}{21.40} & \colorbox{green!15}{32.17} & 21.73

 \\
 \midrule\midrule

\multirow{6}{*}{AttentionInfluence-1.3B w/ LRD }  	& \multirow{6}{*}{1T}  & \multirow{6}{*}{45.26}  &

ARC-C &	 ARC-E &	 ARC(C+E) &	 Wino. 	& Hella. &	 CSQA & OpenBookQA  \\
 & & &  {59.98} 	&  {84.26} 	&  {{72.12}} 	& {68.03} 	 &{75.49} &	  {61.59} & {46.60}  \\
 & & & PIQA& 	 TriviaQA&	 MMLU   &  MMLU-Pro &	AGIEval-en & RACE 	 &DROP 	   \\
&  &   & {79.54} &	 {51.20}	 & {{51.48}} &  {{22.03}} &	 {{26.30}}  &  {42.30} 	&  {36.52}   \\
&  &  & BBH &	 GSM8K& 	 MATH &	 HumanEval & C-Eval & GPQA  \\
& & &  {36.22} &	  {23.73} &	\  {10.80}& 	  {26.55} & 33.06& 24.26	 
 \\
 \midrule\midrule
 
\multirow{6}{*}{AttentionInfluence-7B w/ LRD}  	& \multirow{6}{*}{1T}  & \multirow{6}{*}{\colorbox{green!15}{45.57}}  &   

ARC-C &	 ARC-E &	 ARC(C+E) &	 Wino. 	& Hella. &	 CSQA & OpenBookQA \\
 & & & 56.31& 84.05& 70.18& 67.48& 75.24& \colorbox{green!15}{62.90} & \colorbox{green!15}{47.00}\\
 & & & PIQA& 	 TriviaQA&	 MMLU   &  MMLU-Pro &	AGIEval-en & RACE 	 &DROP 	   \\
 & &  &   \colorbox{green!15}{79.76}& \colorbox{green!15}{51.68}& \colorbox{green!15}{53.18}& 21.70 & \colorbox{green!15}{26.85}& \colorbox{green!15}{42.39}& 36.25  \\
&  & & BBH &	 GSM8K& 	 MATH &	 HumanEval & C-Eval & GPQA \\
& &  &   \colorbox{green!15}{36.80} & \colorbox{green!15}{25.78}& \colorbox{green!15}{11.75}& 25.06& \colorbox{green!15}{36.85} & \colorbox{green!15}{24.87} 

 \\
 
\bottomrule
\end{tabular}%
}
\caption{The ablation results on various benchmarks. The LRD denotes learning rate decay. 
}
\label{tab:abalation_result}
\end{table}

\section{LLM-As-A-Judge Experiment Details}
\label{appd:LLM-As-A-Judge}

We use GPT-4o to evaluate the performance of different data selection methods on the FineWeb-Edu-Dedup domain. On one hand, since most of the data in FineWeb-Edu-Dedup is related to education, we aim for the selected high-quality data to be highly relevant to this domain. Therefore, we design an Education Score based on whether the selected sample content is education-related. On the other hand, we want the selected samples to contain more complex, reasoning-intensive knowledge. Based on this criterion, we design a Reasoning Score.

In summary, we use the following prompt to have GPT-4o score the selected samples:

\begin{methodbox5}{LLM-As-A-Judge}
\subsection*{Prompt:}

Given a piece of text: \textbf{<Selected Sample>}.
Determine whether the text has educational value. If it does, respond with 1; if not, respond with 0.
Then, determine whether the text is reasoning-intensive — that is, whether it contains explicit or implicit logical reasoning chains. If it does, respond with 1; if not, respond with 0.
Respond in the following format:
\begin{verbatim}

\#\#Educational Value Score
<educational value score>

\#\#Reasoning Intensive Score
<reasoning intensive score>
\end{verbatim}
\end{methodbox5}

Although GPT-4o can also be used for scoring pretraining data, different domains require specially designed prompts. Moreover, the computational cost of using GPT-4o for scoring is very high, whereas AttentionInfluence-1.3B has a much lower computational overhead.

\begin{table}[htbp]
\centering
\resizebox{0.99\textwidth}{!}{%
\begin{tabular}{llcccc}
\toprule
Method &  Ranking & Words
\\
\midrule
\multirow{8}{*}{AttentionInfluence} & \multirow{2}{*}
{0\%- 1\%}  &	frac, len, sklearn, append, pyplot, browser, pre, \\
 & & mathbf, 3d, employee, \_\_init\_\_\\
\cmidrule(lr){2-3}
& \multirow{2}{*}{1\%- 10\%}&  well, part, movement, children, appreciation, involve, remember, family	\\
& & growth, treatment, principles, business, b, long, work\\
\cmidrule(lr){2-3}
& \multirow{2}{*}{10\%- 50\%}& 	maximize, paintings, independence, therefore, expenses, regulatory, recall \\ 
& & square, protocols, monitoring, integrity, consistent, channels, inspiring, width\\
\cmidrule(lr){2-3}
& \multirow{2}{*}{50\%- 100\%}	&driver, flying, humble, fourier, smoother, longstanding, owl\\ 
& &  personnel, lawyers, entrenched, beach, brother, oils, wow, desk \\
\hline
\multirow{8}{*}{FineWeb-Edu Classifier} & \multirow{2}{*}{0\%- 1\%} & dimensional, student, 3d, 19th, eco, anti\\  
& &israelite, bmatrix, voter, socio, linspace \\	
\cmidrule(lr){2-3}
& \multirow{2}{*}{1\%- 10\%}& 	creative, based, would, sources, do, system, compared, someone\\
 & &studies, delve, true, turn, only, elements, ultimately\\
 \cmidrule(lr){2-3}
& \multirow{2}{*}{10\%- 50\%}& 	argument, bright, rising, excessive, governments, friendships, complicated, discipline \\
 & & constitutes, hearing, consequences, institutional, match, meets, holocaust\\
 \cmidrule(lr){2-3}
& \multirow{2}{*}{50\%- 100\%}& peek, manifest, reciprocity, obligations, toilet, customized, olive\\
& & validity, enriching, profits, presentations, twelve, originating, arithmetic, nazi
\\
\bottomrule
\end{tabular}%
}
\caption{The high-frequency words of different methods.}
\label{tab:wordo_un_evrlap}
\end{table}

\section{Details of Clustering}
\label{sec: Details of Clustering}
We obtain document embeddings using Sentence-BERT\cite{reimers2019sentence} and apply K-means clustering with $k=100$. For each cluster, we sample representative documents near the cluster center and use GPT-4o to generate descriptive fine-grained (i.e., secondary) category labels, such as \textit{Education–Teaching \& Resources}.

We manually group these secondary labels into six primary categories and report the number of samples falling into each high-level category for both selection methods.


\section{Case Study}
\label{sec:case}

In this section, we present the cases selected by FineWeb-Edu Classifier and AttentionInfluence-1.3B.


\begin{figure}[!tb]
    \centering
    \includegraphics[width=0.99\linewidth]{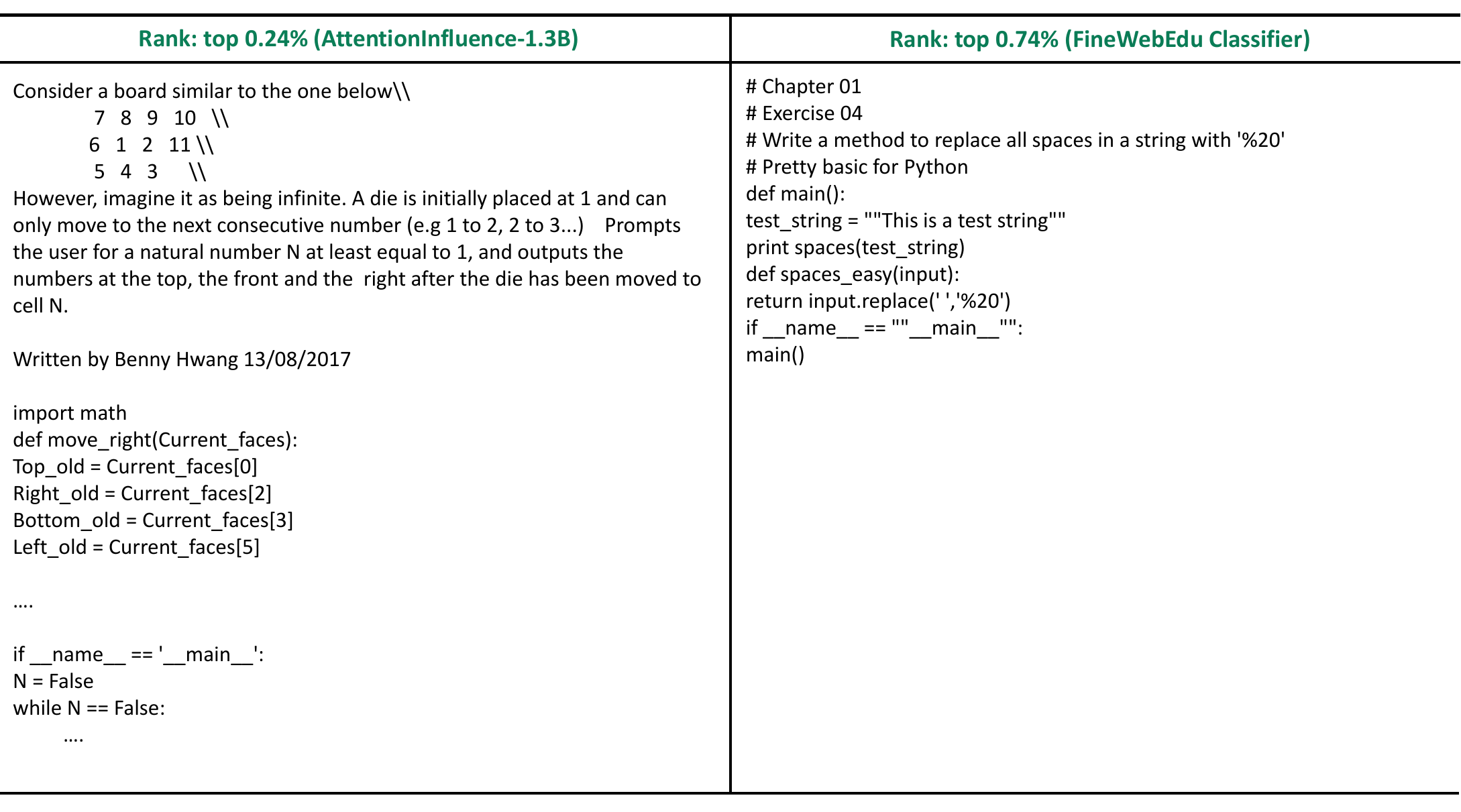}
    \caption{The sample in Python-Edu domain ranked within the top 20\% according to AttentionInfluence-1.3B (\textbf{left}) an FineWeb-Edu Classifier (\textbf{right}).}
    \label{fig:PythonEduCase}
\end{figure}



\begin{figure*}[!tb]
    \centering
    \includegraphics[width=0.99\linewidth]{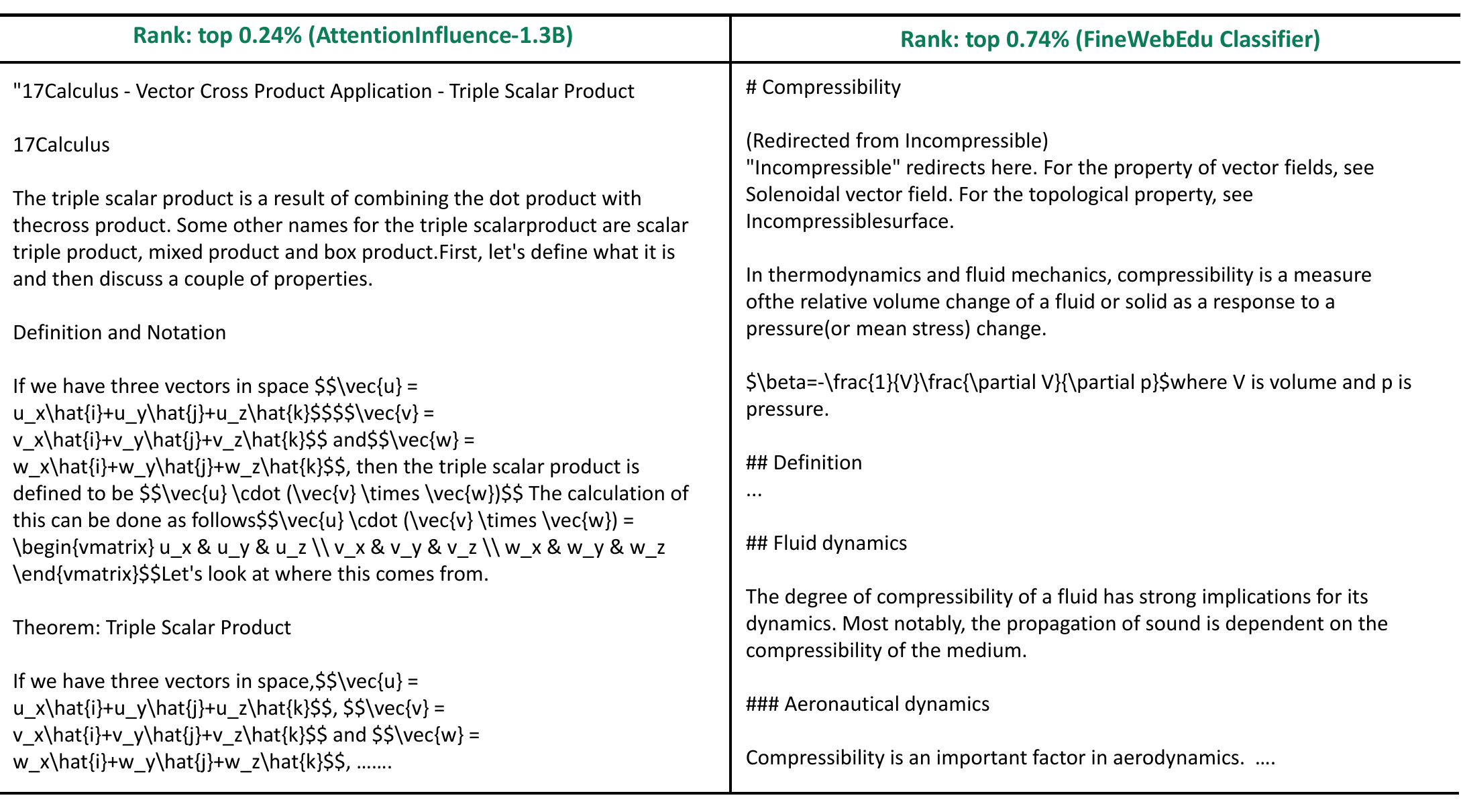}
    \caption{The sample in OpenWebMath domain ranked within the top 20\% according to AttentionInfluence-1.3B (\textbf{left}) and FineWeb-Edu Classifier (\textbf{right}).}
    \label{fig:Open-Web-MathCase}
\end{figure*}

\clearpage

\section{Clustering Case}
\label{Sec:ClusteringCase}
\begin{figure*}[!tb]
    \centering
    \includegraphics[width=0.99\linewidth]{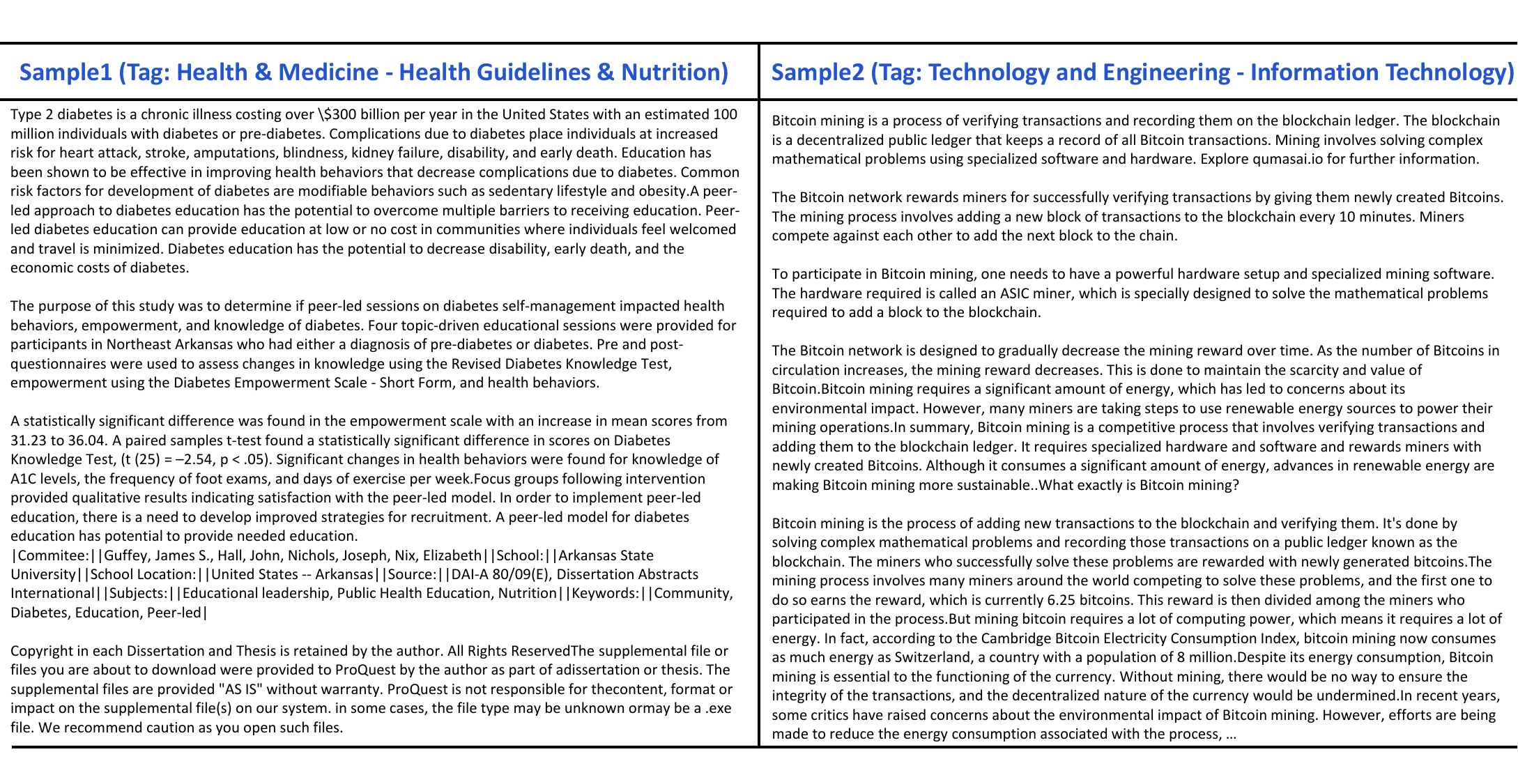}
    \caption{The samples of a clustering in data in the Cosmopedia-V2 domain ranked within 20\% according to AttentionInfluence.}
    \label{fig:Clustering_case}
\end{figure*}

As shown in \autoref{fig:Clustering_case}, we present the two clustering cases in the Cosmopedia-V2 domain.

\clearpage
\section{Cases of AttentionInfluence based on 1.3B and 7B Models}

\begin{figure}[!tb]
    \centering
    \includegraphics[width=0.99\linewidth]{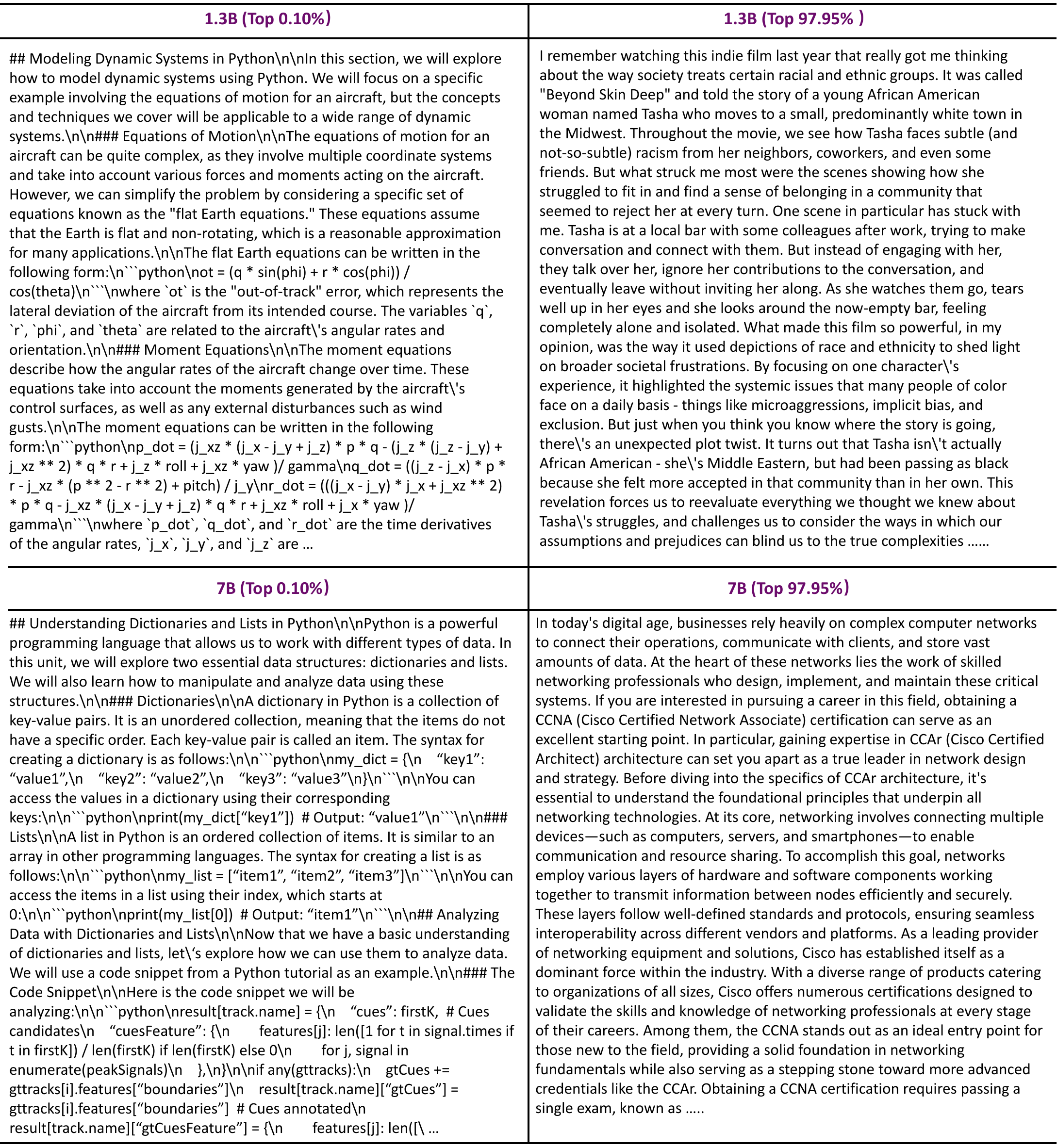}
    \caption{The cases of AttentionInfluence in Cosmopeida-V2 domain.}
    \label{fig:Cosmopeida-V2}
\end{figure}

\begin{figure}[!tb]
    \centering
    \includegraphics[width=0.99\linewidth]{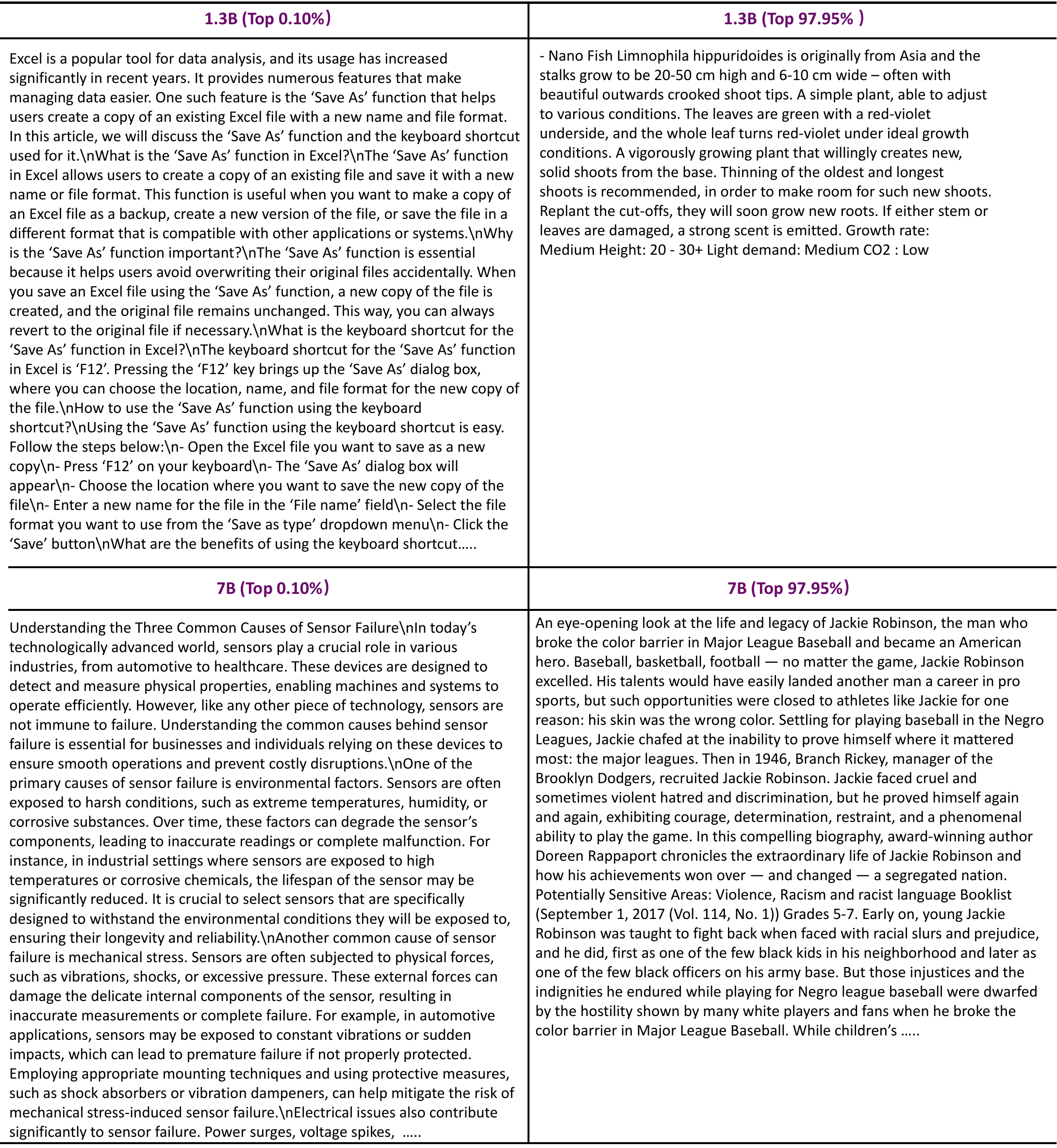}
    \caption{The cases of AttentionInfluence in FineWeb-Edu-Dedup domain.}
    \label{fig:FineWeb-Edu}
\end{figure}

\begin{figure}[!tb]
    \centering
    \includegraphics[width=0.99\linewidth]{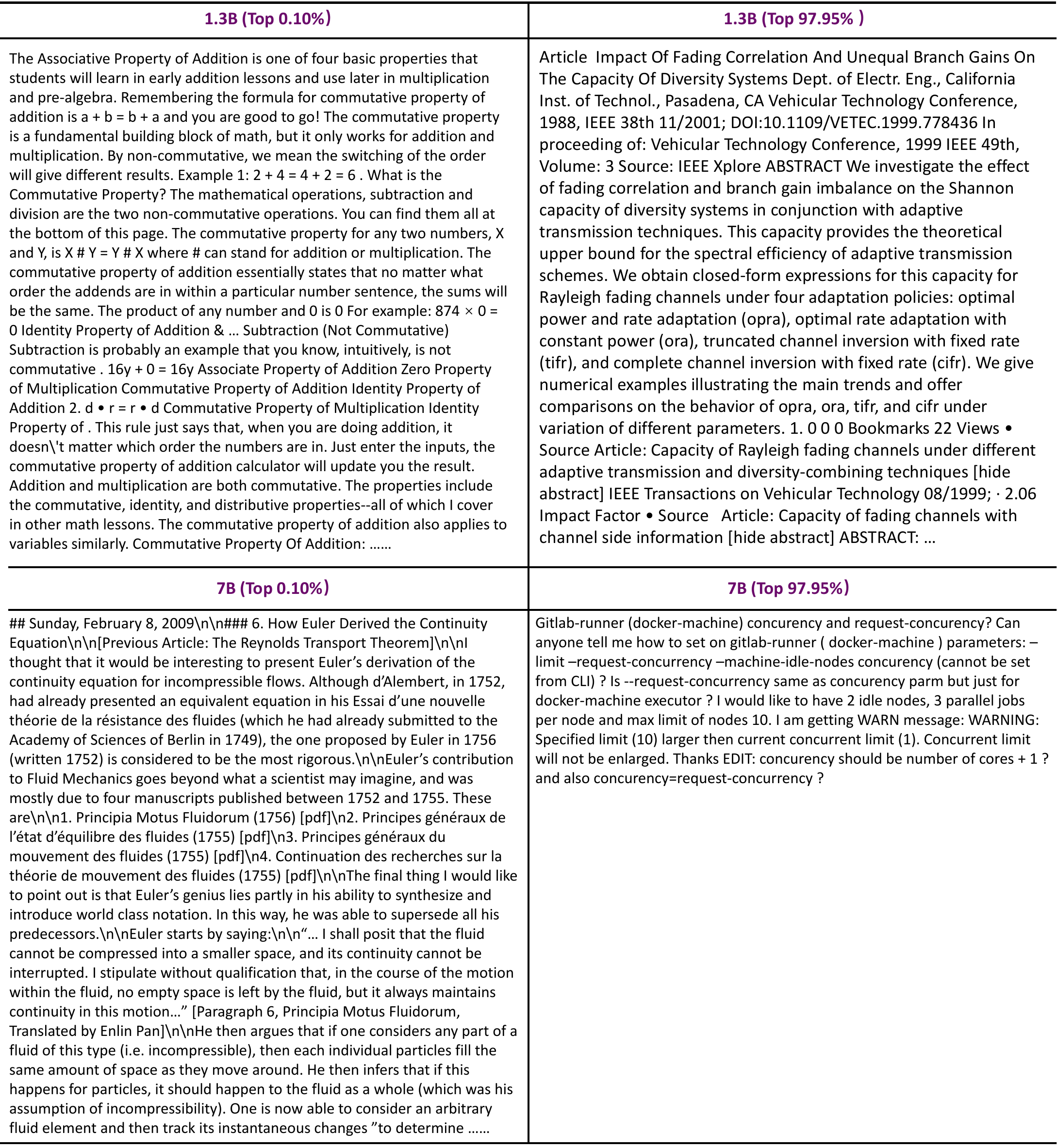}
    \caption{The cases of AttentionInfluence in OpenWebMath domain.}
    \label{fig:Open-Web-Math}
\end{figure}

\begin{figure}[!tb]
    \centering
    \includegraphics[width=0.99\linewidth]{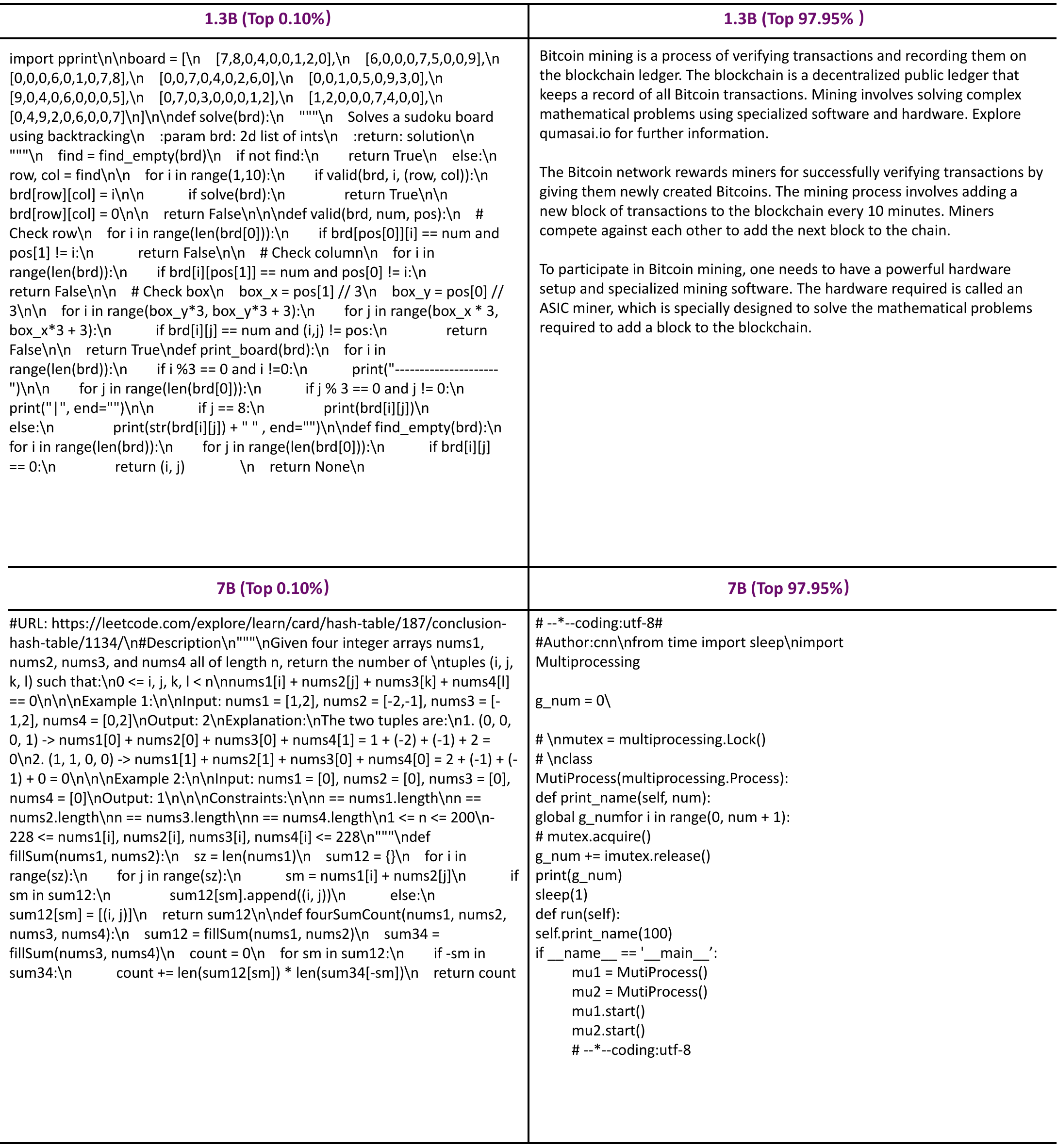}
    \caption{The cases of AttentionInfluence in Python-Edu domain.}
    \label{fig:Python-Edu}
\end{figure}

As shown in \autoref{fig:Cosmopeida-V2}, \autoref{fig:FineWeb-Edu}, \autoref{fig:Open-Web-Math}, and \autoref{fig:Python-Edu}, we present some cases with different score levels.

\clearpage

\section{High Frequency Words}
\label{appendix: High Frequency Words}

\begin{figure*}[!tb]
    \centering
    \includegraphics[width=0.9\linewidth]{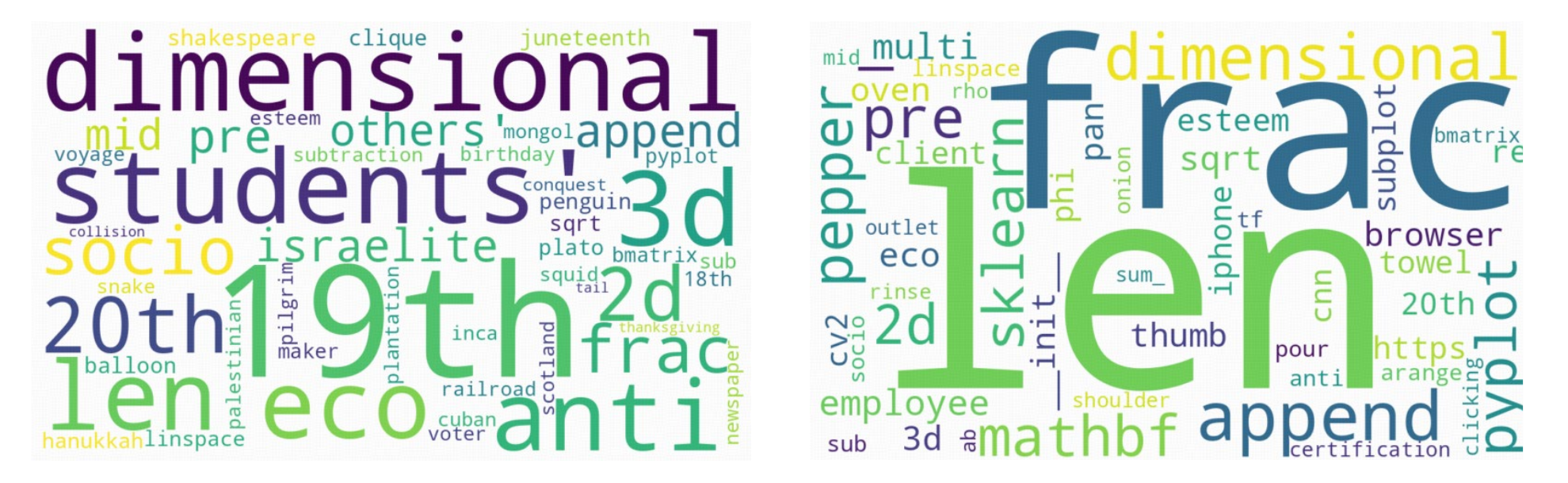}
    \caption{The cloud maps of the data selected by AttentionInfluence and FineWeb-Edu Classifier, respectively. The left part is the cloud map of FineWeb-Edu Classifier, the right part is that of AttentionInfluence.}
    \label{fig:cloud_map}
\end{figure*}

As illustrated in \autoref{fig:cloud_map}, we visualize the respective word clouds of AttentionInfluence-1.3B and the FineWeb-Edu Classifier after removing overlapping high-frequency words in the Cosmopeida-V2 domain. The resulting word clouds clearly highlight their distinct focal points, indicating a complementary relationship between the two models. To gain deeper insights, we further examine representative samples corresponding to the key terms in each word cloud.


\begin{figure*}[!tb]
    \centering
    \includegraphics[width=0.99\linewidth]{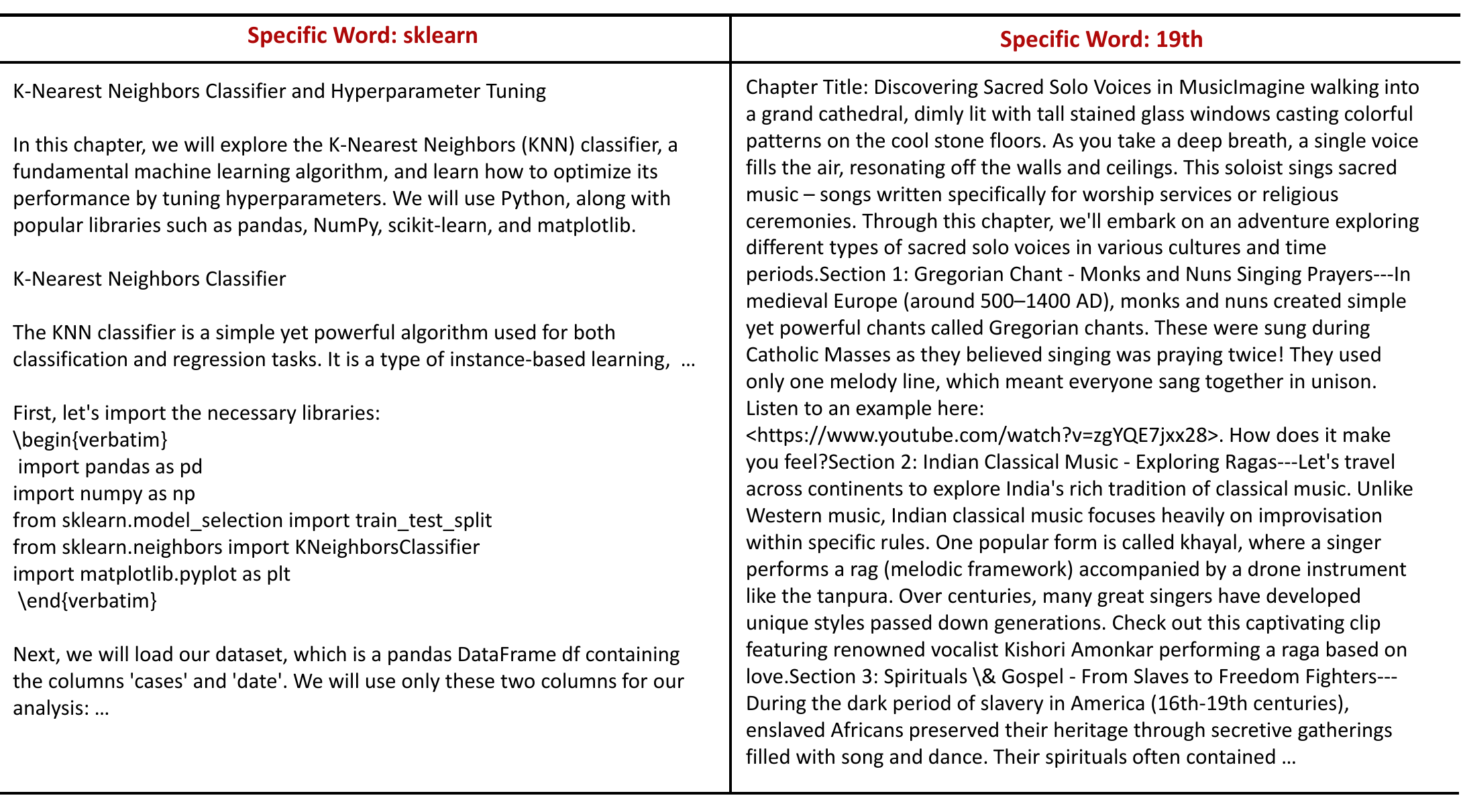}
    \caption{The sample of a doc containing the specific word selected by AttentionInfluence-1.3B (\textbf{left}) and FineWeb-Edu Classifier (\textbf{right}).}
    \label{fig:SpecialWord}
\end{figure*}

\clearpage
\section{Limitations and Opportunities}
While our experimental results demonstrate the effectiveness of AttentionInfluence, several important aspects warrant further investigation. We identify five key areas for future research:
\begin{itemize}[leftmargin=1.5em]
    \item Our current experiments demonstrate the effectiveness of AttentionInfluence up to 7B parameters and 1,000B tokens of training budget. Extending this approach to long-horizon training and larger-scale models requires a highly expensive computational cost, and we leave it for future research.
    \item Due to limited manpower, we do not investigate the effects of selected data by AttentionInfluence on the final performance of models, followed by post-training. We hypothesize that reinforcement learning will amplify the good effects of selected data by AttentionInfluence.
    \item While this work focuses on selecting data from short texts, AttentionInfluence can be readily extended to long texts to identify high-quality samples characterized by long-range dependencies.
    \item We conduct experiments with alternative approaches for identifying important attention heads, such as the method proposed by \citet{fu2024not}, which produces a partially overlapping yet distinct set of heads compared to ours. 
    Training LLMs based on the data selected by these heads can achieve comparable downstream evaluation performance.
    More recently, \citet{zhu2025focus} introduces another compatible method that can be incorporated into our framework.

These results demonstrate that AttentionInfluence serves as a flexible and general framework: by defining an appropriate proxy task, one can identify task-relevant attention heads and select associated data via masking. The entire pipeline operates without any supervision signals and is modular by design, allowing the proxy task to be easily replaced depending on the target domain or task. Moreover, the framework is effective even when applied to small pretrained language models, making it practical and scalable for a wide range of data selection scenarios.

More comprehensive proxy tasks can also be designed to better capture specific types of data within the AttentionInfluence framework, further expanding its applicability and customization potential.

    \item The combined effect of multiple heads remains unknown. Moreover, this work does not involve research on the MLP. Substantially more in-depth research endeavors are required to unearth the more fundamental and intrinsic mechanisms underpinning language models.
\end{itemize}

\end{document}